\documentclass{article}
\pdfoutput=1

    \usepackage[final,nonatbib]{neurips_2023}

\newcommand{\rv}{RETVec\xspace}
\newcommand{\rvr}{RETVec-raw\xspace}
\newcommand{\rvb}{RETVec\xspace}

\usepackage{todonotes}
\usepackage{caption}
\usepackage{graphicx}
\usepackage{subfigure}
\usepackage{booktabs} % for professional tables
\usepackage{xspace}
\usepackage{array}
\usepackage{multirow}
\usepackage{ulem}

\usepackage[utf8]{inputenc} % allow utf-8 input
\usepackage[T1]{fontenc}    % use 8-bit T1 fonts
\usepackage{hyperref}       % hyperlinks
\usepackage{url}            % simple URL typesetting
\usepackage{booktabs}       % professional-quality tables
\usepackage{amsfonts}       % blackboard math symbols
\usepackage{nicefrac}       % compact symbols for 1/2, etc.
\usepackage{microtype}      % microtypography
\usepackage{xcolor}         % colors

\title{RETVec: Resilient and Efficient Text Vectorizer}

\author{
  Elie Bursztein \\
  Google \\
  \texttt{elieb@google.com} \\
  \And
  Marina Zhang \\
  Google \\
  \texttt{marinazh@google.com} \\
  \And
  Owen Vallis \\
  Google \\
  \texttt{ovallis@google.com} \\
  \And
  Xinyu Jia \\
  Google \\
  \texttt{jiaxinyu@google.com} \\
    \And
   Alexey Kurakin \\
  Google \\
  \texttt{kurakin@google.com} \\
}

\begin{document}

\maketitle

\begin{abstract}
This paper describes \rv, an efficient, resilient, and multilingual text vectorizer designed for neural-based text processing. \rv combines a novel character encoding with an optional small embedding model to embed words into a 256-dimensional vector space. The \rv embedding model is pre-trained using pair-wise metric learning to be robust against typos and character-level adversarial attacks. In this paper, we evaluate and compare \rv to state-of-the-art vectorizers and word embeddings on popular model architectures and datasets. These comparisons demonstrate that \rv leads to competitive, multilingual models that are significantly more resilient to typos and adversarial text attacks.
\rv is available under the Apache 2 license at \url{https://github.com/google-research/retvec}.
\end{abstract}

\section{Introduction}
\label{sec:int}

Researchers have proposed many techniques for converting text into dense representations suitable for training neural networks, from the early use of character bi-grams and tri-grams to more modern semantic word embeddings and state-of-the-art subword vectorizers. Each of these approaches aims to mitigate issues caused by out-of-vocabulary (OOV) tokens, including intentional text perturbations such as adversarial attacks and unintentional ones such as typos. These approaches leverage techniques such as subword-level tokenization~\cite{kudoSentencePieceSimpleLanguage2018} and decomposing unknown words into n-grams~\cite{bojanowskiEnrichingWordVectors2017}.

However, we find that each of these approaches suffers from at least one of following drawbacks:
\begin{itemize}
     \setlength{\itemsep}{0pt}
      \setlength{\parskip}{0pt}
  \setlength{\parsep}{0pt}
    \item They have insufficient resilience against typos and adversarial attacks~\cite{morrisTextAttackFrameworkAdversarial2020}.
    \item They require large dictionaries and embedding lookup tables.
    \item They exhibit poor performance on certain languages or in multilingual settings.
\end{itemize}

\rv addresses these issues by combining a novel UTF-8 character encoder with an optional small model (230k parameters). This model projects encoded words into a 256 dimensional \textbf{syntactic metric embedding} as detailed in Section~\ref{sec:arc}. \rv embeddings are trained using pair-wise metric learning~\cite{wangMultiSimilarityLossGeneral2020}, ensuring that words containing typos are embedded close to the the original word.

% Models trained with \rv are competitive and significantly improve a model's resilience against both typos and adversarial attacks compared to other popular vectorizers and word embeddings.

% Because \rv works directly on UTF-8 characters, it does not require any dataset pre-processing and has no OOV tokens. 

\rv does not require dataset pre-processing and does not have OOV tokens because it accepts all valid UTF-8 characters. \rv's embedding model is trained on a word dataset with more than 157 languages. \rv is also space-efficient (<1MB) since it does not require a large embedding lookup table. This reduces model memory footprints, making \rv-based models ideal candidates for on-device model deployment where storage, bandwidth, and memory resources are scarce. Through a series of extensive experiments on different model architectures, and on a wide range of datasets, we demonstrate that \rv outperforms or is comparable to state-of-the-art vectorizers and word embeddings including BPE~\cite{sennrichNeuralMachineTranslation2016}, SentencePiece~\cite{kudoSentencePieceSimpleLanguage2018}, and fastText~\cite{bojanowskiEnrichingWordVectors2017}. Overall, \rv outperforms other vectorizers on text classification tasks by about 1\%, while being up to 15\% more resilient to typos at 20\% word typo rate, and less susceptible to character-level adversarial attacks by over 10\%.

Our paper makes the following contributions:

\begin{itemize}
     \setlength{\itemsep}{0pt}
      \setlength{\parskip}{0pt}
  \setlength{\parsep}{0pt}
    \item We introduce \rv, a resilient, efficient, and multilingual text vectorizer designed for neural-based text processing. 
    
    % \rv consists of a novel character encoder and a typo-robust word embedding, pre-trained using pair-based deep metric learning.

    % \rv works seamlessly on every combination of languages without the need to adapt or retrain it.
    % Removed the following as this doesn't apply to non-whitespace charsets like Japanese, Chinese, and Korean
    % \item the \rv embedding works well on top of simple vectorizer splitting text on whitespaces and punctuation, which makes the vectorization process much faster compared to other modern vectorizers. In particular, in section~\ref{sec:spe} we show that \rv with simple vectorizer is as fast or faster than other vectorizers on multi-core CPU or when a GPU is available as it exploits TensorFlow vectorization efficiently. \rv also uses significantly less memory than the other vectorizers as it is stateless. On a single-core CPU using the optional model is about 4x slower than SentencePiece.
    
    \item  We show that \rv is faster and less memory intensive than other vectorizers on multi-core CPUs and on GPUs.

    \item We demonstrate that models trained with \rv have slightly higher accuracy, greater resilience to typos, and significantly better resilience to adversarial attacks compared to models trained with the other vectorizers.

    \item We provide a TensorFlow implementation of \rv, including its pre-trained models, alongside the code to reproduce our benchmarks under the Apache 2 license at \\ \url{https://github.com/google-research/retvec}.
\end{itemize}

% Additionally, throughout the architecture~\ref{?}, evaluation sections\ref{sec:cla, sec:res, sec:adv}, and the ablation study section~\ref{?} we demonstrate empirically that using a larger and slower transformer model as \rv's optional model does not always lead to better results despite having a significantly lower loss. Similarly, this larger model also exhibits slightly lower resilience to adversarial attacks than our base model (sec~\cite{sec:adv}).

\section{Background}
\label{sec:bac}
% Deep neural networks (DNNs) operate on floating-point tensors. To process textual information, DNNs have to convert text into a floating-point representation. This is typically achieved by splitting text into tokens and encoding each of those tokens into a dense floating-point vector using a vectorizer. 

% in the following way:
% \setlength{\parskip}{0pt}
% \setlength{\parsep}{0pt}
% \begin{itemize}
%     \setlength{\itemsep}{0pt}
%     \setlength{\parskip}{0pt}
%     \setlength{\parsep}{0pt}
%     \item First, a tokenizer splits text into tokens. These could be words, pieces of words or even individual characters.
%     \item  Then, each token is encoded as a floating point vector using an embedding algorithm.
% \end{itemize}

\paragraph{Text Vectorizers}
Text vectorizers play a major role in deep neural networks' (DNN) performance by ensuring that the input text is properly segmented into tokens (typically words or sub-words) and embedded into a dense, float-point representation for the model to use. The traditional approach for text vectorization splits texts into words using whitespace and punctuation and uses a word embedding lookup table to map each word into a dense vector. The embedding lookup table can be constructed using algorithms such as Word2Vec~\cite{mikolovEfficientEstimationWord2013}, GloVe~\cite{penningtonGloVeGlobalVectors2014} and fastText~\cite{bojanowskiEnrichingWordVectors2017}, or by randomly initializing an embedding table and training it as a part of the LLM pre-training process (like GPT-2~\cite{BetterLanguageModels2019}). Embedding lookup tables are typically large and limited to known, in-vocabulary tokens which significantly reduces their representational capabilities in the presence of typos and adversarial attacks.

Subword tokenizers such as WordPiece~\cite{devlinBERTPretrainingDeep2019}, BPE~\cite{sennrichNeuralMachineTranslation2016} and SentencePiece~\cite{kudoSentencePieceSimpleLanguage2018}) split text into subwords to reduce vocabulary size and mitigate OOV issues.
%by representing unknown words as combinations of known tokens. SentencePiece deals better with segmentation and OOV tokens, and is used by most state-of-the-art models these days~\cite{raffelExploringLimitsTransfer2020,chowdheryPaLMScalingLanguage2022}.
SentencePiece is used by many state-of-the-art text models~\cite{raffelExploringLimitsTransfer2020,chowdheryPaLMScalingLanguage2022}. All of these vectorizers require a separate training phase to adapt to the targeted dataset and require shipping extra files with the models to be used at inference time.

\paragraph{Pair-based learning}
The \rv model is trained using pair-based metric learning.
Pair-based learning aims to learn an embedding space, where embedded vectors of similar samples are encouraged to be closer together while dissimilar ones are pushed apart from each other~\cite{xingDistanceMetricLearning}. This approach has been successfully applied to various tasks including image retrieval~\cite{heTripletCenterLossMultiView2018, wangMultiSimilarityLossGeneral2020, sunCircleLossUnified2020}, face recognition~\cite{schroffFaceNetUnifiedEmbedding2015}, and zero-shot learning~\cite{zhangZeroShotLearningJoint2016}.

% In the deep learning setting, this type of learning is accomplished using loss functions that minimize/maximize the pairwise distances within the embedding space. In this paper we experiment with {\tt Multi Similarity loss}~\cite{wangMultiSimilarityLossGeneral2020} and {\tt Circle loss}~\cite{sunCircleLossUnified2020}, two of the most recent and most efficient pairwise losses.

% As discussed in detail in section~\ref{sec:pretrain}, we use a pairwise loss to pre-train \rv optional models by pulling words close to their typo-laden version and away from other words. This training objective is empirically the best-suited for increasing the resilience of RetVec embeddings against typographical augmentations and to reduce OOV issues such as those evaluated in section~\ref{sec:abl}.

\paragraph{Typos and Adversarial Attacks}
One challenge faced by text vectorizers is the existence of typographical errors that might not appear in the training dataset. At inference time, these typographical errors can lead the vectorizer to incorrectly output OOV ids for valid words. This behavior ultimately leads to lower accuracy, particularly for retrieval tasks where human queries are known to exhibit between 16\% and 19\% typos rate~\cite{zhuangDealingTyposBERTbased2021, hagenLargeScaleQuerySpelling2017}. In many anti-abuse settings such as spam classification, the typo rate used by attackers in attempt to evade defense systems is much higher~\cite{DADA2019e01802}. It is also well-known that text models are vulnerable to different types of adversarial attacks~\cite{morrisTextAttackFrameworkAdversarial2020}. Many character-level adversarial attacks, which can be viewed as the most severe form of intentional typos, have been proposed in the literature (~\cite{liTextBuggerGeneratingAdversarial2019}, ~\cite{gaoBlackboxGenerationAdversarial2018}, ~\cite{pruthiCombatingAdversarialMisspellings2019}).

%Query datasets compiled from real-world logs exhibits a typographical error rate between 16\% and 19\%~\cite{hagenLargeScaleQuerySpelling2017}. In adversarial settings such as classifying spam the amount of typo used by attackers in an attempt to evade classifier is much higher.\todo{cite some spam classifier paper}.

% Furthermore, text models are vulnerable to adversarial attacks that intentionally corrupt words to force the model to predict wrongly. Many character-level adversarial attacks aim to maximize the loss of accuracy per word manipulation by triggering the most detrimental OOV cases. In recent years, many successful adversarial text attacks
% ~\cite{morrisTextAttackFrameworkAdversarial2020} have been developed, showcasing the vulnerability of text models to these types of augmentations. These attacks can be viewed as the most severe form of intentional typos.

% \todo{write something about anti abuse and increased typo percent? like in spam detection, adversaries will often insert gibberish beyond what is seen in normal human typos and misspellings}

% \paragraph{Adversarial Attacks and Defenses}

\section{RETVec}
\label{sec:arc}
\label{ref:retvec-overview}

In this section, we describe how \rv works. We start off by introducing its novel character encoding scheme, detail the design of its model, and finally describe its pre-training procedure.

\begin{figure}[h]
\centering{
\includegraphics[width=\textwidth]{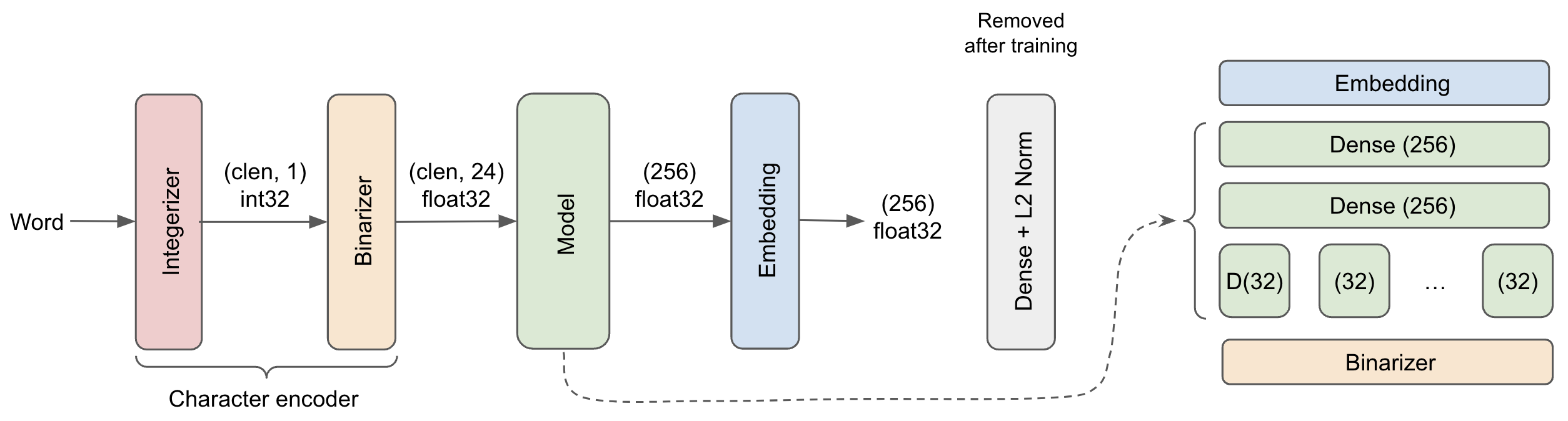}
\caption{\rv architecture overview - the output shape of each layer is in parenthesis. The \texttt{clen} indicates the number of characters used per word - 16 characters by default. The batch and word sequence length dimensions are omitted.}
\label{fig:arch}}
\end{figure}

\subsection{\rv Character Encoder}

The \rv character encoder, as shown in Figure~\ref{fig:arch}, is composed of two layers:  an {{\tt Integerizer} layer} and a {\tt Binarizer} layer. The character encoder enables \rv to encode all UTF-8 characters in an efficient and stateless manner. Encoding a word is achieved by first converting its characters into their UTF-8 codepoints ({\tt Integerizer}) and then converting this integer representation into a flattened, binary (little-endian) representation ({\tt Binarizer}). The {\tt Binarizer} uses a compact 24-bit binary character representation to encode all valid UTF-8 characters.
Additionally, \rv outputs a fixed-length, dense word embedding using a maximum of 16 characters per word. In the ablation study (Section~\ref{sec:abl}), we empirically find that 16 characters per word works well and increasing the max word length provides no further accuracy gains.

The {\tt Integerizer} is similar to the one used in ByT5~\cite{xueByT5TokenfreeFuture2022} which also encodes characters into their UTF-8 codepoints. What makes the \rv encoder unique is that the codepoints are first converted to their binary representation and then flattened to create the word representation. 
This novel binary representation is both easy to learn and significantly more compact than a one-hot representation.
We tested many other approaches to convert the {\tt Integerizer} output into a more compact form, including the use of a single float per character, but these more compact representations did not appear to be learnable.
Additionally, a key design choice is to keep the {\tt Integerizer} and the {\tt Binarizer} separated for efficiency reasons: when using a remote accelerator, such as a TPU, it is more efficient to send the output of the {\tt Integerizer} ($16 \times$ int32) across the network rather than directly sending the binarization representation ($16 \times 24$ float32)).

% We note that while TensorFlow doesn’t have native support for converting integers to binary, we found out that, as reported in the performance evaluation in Section~\ref{sec:spe}, writing it using bit shifts with a precomputed matrix to be sufficiently fast to not be a bottleneck. If \rv becomes widely used, adding binarization native support in deep-learning frameworks will improve its speed further.

\subsection{\rv Model Architecture}

While the \rv\space{\tt Binarizer} provides an efficient word representation, it is not competitive with state-of-the-art vectorizers.
To address this, we add a small model on top of the {\tt Binarizer} output. This improves accuracy and enables \rv to outperform other vectorizers (see Section~\ref{sec:cla}). Additionally, this reduces the embedding size from 384 to 256 \texttt{float32}s and further improves resilience to adversarial attacks (see Section~\ref{sec:adv}). 
The performance gains only incur a small increase in computational cost, making the use of the pre-trained model a worthwhile trade-off. To better understand what gains can be attributed to the character encoder, the remainder of the paper reports the performance for ``\rv with the model'' as \rv, and to understand what can be attributed to the model we report ``\rv without the model'' as \rvr.\\

The \rv model, as visible in Figure~\ref{fig:arch}, is composed of a dense projection layer, a flatten layer, a dense compression layer, and the embedding layer. The embedding layer is a dense layer with a {\tt tanh} activation to scale the embedding values between 0 and 1, while the rest of the network uses {\tt gelu} activations. Increasing capacity by adding additional layers or using more powerful architectures did not improve performance on our benchmarks (Section~\ref{sec:abl}). Detailed ablation studies on loss hyperparameters, dropout rates, activation functions, and model capacity can be found in Appendix~\ref{app:abl}.

\subsection{\rv Pre-training Procedure}
\label{sec:pretrain}
\paragraph{Dataset}
The \rv model is pre-trained on a typo-augmented version of the 157-language fastText words datasets \cite{graveLearningWordVectors2018}. The original words are extracted from the Common Crawl corpus. Our dataset then takes the set of words over the union of all 157 languages, yielding 88.8M unique tokens.
To help with generalization, we supplement this dataset by adding 8.88M (10\%) randomly generated tokens, where the length of a token is randomized between 1 and 16. Each token consists of a random set of UTF-8 characters.
Finally, we create 20 versions of each token (16 augmented and 4 non-augmented versions) such that ~80\% of the tokens in the training set are typo-augmented. At the end of the generation process we end-up with a combined 1.9B token dataset.

\paragraph{Augmentations}
Token augmentation consists of randomly inserting up to 4 typos per token up to 25\% of the token length. This is consistent with an observed maximum human error frequency of around 20\%  \cite{hagenLargeScaleQuerySpelling2017}. We use 22 distinct typo augmentations, which can be grouped into four categories: deletion, insertion, substitution, and transposition. For each token, we randomly select a target augmentation percentage between 0-25\%, and for each augmentation step we randomly apply an augmentation from one of the four typo categories. The full list of augmentations used is reported in Appendix~\ref{app:aug}.

\paragraph{Training}
\rv model is trained using pair-wise learning and uses the Multi-Similarity Loss~\cite{wangMultiSimilarityLossGeneral2020} ($\alpha = 4$, $\beta = 40$, $\lambda = 0.5$, and $\epsilon = 0.1$). We construct example pairs by creating batches that always contain two variations of the same word. The model is trained for 500k steps with batch size = 1024, using Adam with max learning rate = 0.001, $\beta_1=0.9$, $\beta_2=0.999$, and cosine decaying the learning rate to 0.0001 during training. Full training hyperparameters can be found in Appendix~\ref{app:tra}. We experimented with additional losses and other self-supervised tasks, reported in Appendix~\ref{app:abl}, but none of them yielded better results.

\begin{figure}[h]
% \vspace{-3mm}
\centering{
\includegraphics[width=\textwidth]{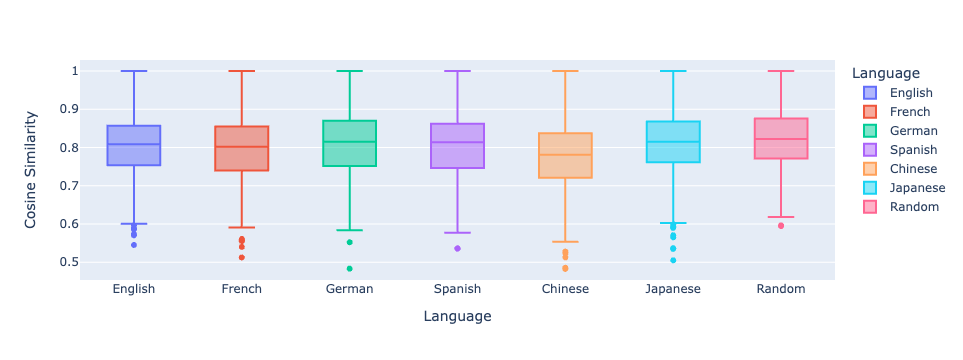}
\caption{The cosine similarity distributions of \rv embeddings for 1000 pairs of augmented and non-augmented versions of words, selected languages shown. `Random` language refers to randomly-generated UTF-8 strings.}
% \vspace{-3mm}
\label{fig:box_plot}}
\end{figure}

\paragraph{Embedding Quality} To visualize and evaluate the training quality and embedding consistency across languages, we randomly sample pairs of augmented and non-augmented words from select languages and compare the distribution of their cosine similarities. As shown in Figure~\ref{fig:box_plot}, we observe that the mean and variance of the cosine similarity distributions are fairly consistent across various languages, which suggests that the model has learned near-uniformly across languages. The mean cosine similarity between word pairs in Chinese is slightly lower, which could be attributed to the fact that the average length of words is significantly shorter in Chinese. The distribution for random strings is also consistent with other languages' distributions, suggesting that \rv is able to generalize to previously unseen words and embed words with rare UTF-8 characters meaningfully, which would greatly contribute towards its adversarial resilience. In Appendix~\ref{app:emb_viz}, we used the TensorBoard Embedding Projector to visualize \rv embeddings in a 3D space and demonstrate that syntactically-similar words are clustered together, adding further evidence that \rv is meaningfully projecting tokens into the embedding space.

% We experimented with various types of losses, including: reconstruction loss such as MSE to reconstruct the original word; contrastive loss such as SimSiam~\cite{chenExploringSimpleSiamese2020}; token replacement detection and pair-based losses like Multi-Similarity Loss~\cite{wangMultiSimilarityLossGeneral2020} and Circle Loss~\cite{sunCircleLossUnified2020a}. Overall, as reported in appendix~\ref{app:loss}, we found that pair-based losses work best and combining them with other losses does not empirically bring any improvement for neither \rvb nor \rvl.

\section{Evaluation: Speed}
\label{sec:spe}

\begin{table*}[t]
  \resizebox{\textwidth}{!}{%
    \begin{tabular}{|l|c|c|c|c|c|ccc|ccc|}
      \hline
      & \multicolumn{1}{l|}{} & \multicolumn{1}{l|}{} & \multicolumn{1}{l|}{} & \multicolumn{1}{l|}{} & \multicolumn{1}{l|}{} & \multicolumn{3}{c|}{CPU} & \multicolumn{3}{c|}{GPU} \\
      Name          & \multicolumn{1}{l|}{\begin{tabular}[c]{@{}l@{}}Embedding \\ Size\end{tabular}} & \multicolumn{1}{l|}{\begin{tabular}[c]{@{}l@{}}Vocabulary \\ Size\end{tabular}} & \multicolumn{1}{l|}{\begin{tabular}[c]{@{}l@{}}Preprocessing \\ Time\end{tabular}} & \multicolumn{1}{l|}{\begin{tabular}[c]{@{}l@{}}System \\ Memory (GB)\end{tabular}} & \multicolumn{1}{l|}{\begin{tabular}[c]{@{}l@{}}Model \\ Params\end{tabular}} & \multicolumn{1}{l|}{Wall time (s)} & \multicolumn{1}{l|}{Usage}  & \multicolumn{1}{l|}{\begin{tabular}[c]{@{}l@{}}Core \\ Sec\end{tabular}} & \multicolumn{1}{l|}{Wall (s)} & \multicolumn{1}{l|}{Usage} & \multicolumn{1}{l|}{Memory (GB)} \\ \hline
      SentencePiece & 256                                            & 32k                                            & 749                                            & 10.9                                           & 8M                                             & \multicolumn{1}{c|}{877}     & \multicolumn{1}{c|}{124\%}  & 1087                                           & \multicolumn{1}{c|}{870}       & \multicolumn{1}{c|}{100\%}     & 0.5                               \\ \hline
      BPE           & 256                                            & 32k                                            & 861                                            & 3.6                                            & 8M                                             & \multicolumn{1}{c|}{1062}    & \multicolumn{1}{c|}{169\%}  & 1795                                           & \multicolumn{1}{c|}{1050}       & \multicolumn{1}{c|}{100\%}     & 0.5                               \\ \hline
      FastText      & 300                                            & -                                              & 420                                            & 43.9                                           & 0                                              & \multicolumn{1}{c|}{845}     & \multicolumn{1}{c|}{52\%}   & 439                                            & \multicolumn{1}{c|}{-}       & \multicolumn{1}{c|}{-}     & -                               \\ \hline
      Whitespace   & 256                                            & 32k                                            & 132                                            & 1.7                                            & 8M                                             & \multicolumn{1}{c|}{496}     & \multicolumn{1}{c|}{126\%}  & 625                                            & \multicolumn{1}{c|}{396}     & \multicolumn{1}{c|}{100\%} & 0.5                             \\ \hline
      \rvr          & 384                                            & -                                              & 0                                              & 1.9                                  & 0                                             & \multicolumn{1}{c|}{225}     & \multicolumn{1}{c|}{451\%}  & 1015                                           & \multicolumn{1}{c|}{235}     & \multicolumn{1}{c|}{20\%}  & 0.4                             \\ \hline
      \rvb          & 256                                            & -                                              & 0                                              & 2.1                                    & 230K                                           & \multicolumn{1}{c|}{570}     & \multicolumn{1}{c|}{840\%}  & 4789                                           & \multicolumn{1}{c|}{290}     & \multicolumn{1}{c|}{40\%}  & 0.5                             \\ \hline
    \end{tabular}%
  }
  \caption{Comparison of speed, preprocessing time, memory usage, and CPU/GPU usage for all benchmarked vectorizers when vectorizing the Multilingual Amazon Reviews training dataset.}
  \label{tab:tok}
\end{table*}

In this section, we evaluate how quickly \rv can process and vectorize datasets compared to the other commonly used vectorizers. We considered SentencePiece \cite{kudoSentencePieceSimpleLanguage2018}, BPE \cite{sennrichNeuralMachineTranslation2016}, a simple whitespace vectorizer, and fastText \cite{bojanowskiEnrichingWordVectors2017}. We use the HuggingFace implementation of BPE and the official SentencePiece implementation.

\paragraph{Setup}
We use the Multilingual Amazon Reviews~\cite{mcauleyHiddenFactorsHidden2013} dataset for the vectorization speed evaluations. The corpus is composed of 1.2M tokens -- 200k tokens for 6 languages (English, French, German, Spanish, Chinese, Japanese). All the experiments are run on a standard Google Cloud VM with 16 CPU cores and a V100 Nvidia GPU. We disable the GPU visibility to perform the CPU measurements and use the {\tt GNU time} command to measure the total wall time, CPU usage, and system memory. We use {\tt nvidia-smi} to record GPU usage and memory. We report both the time it takes to adapt to the datasets and how long it takes to vectorize the entire dataset end-to-end, with and without GPU acceleration. We also report normalized CPU-core times as \rv (and other vectorizers to a lesser extent) benefits from multi-threaded acceleration. Finally, we report memory usage for both system and GPU, since they are important metrics when developing models for memory-constrained devices such as IoT devices and smartphones.

% by running it at 10 second intervals during execution. Table ~\ref{tab:tok} reports a single number for the GPU usage as we discovered that GPU usage and memory remains stable throughout the runs.

\paragraph{Results}
We found that \rvr and \rvb are the fastest vectorizers when a GPU is available and are in the top three when multi-core CPU are available (Table~\ref{tab:tok}), which is common on current devices including smartphones. \rvb's performance on CPU stems from its ability to efficiently use the available cores, as visible in the Core Sec column – we expect that more optimized versions of SentencePiece and BPE could achieve similar performance.

\section{Evaluation: Classification}
\label{sec:clase}
\label{sec:cla}

In this section, we compare the classification training performance of \rv against other vectorizors and word embeddings. We train the models from scratch and report classification accuracy on a wide range of datasets to benchmark the generality of the approach.

\paragraph{Setup}

% \begin{table}[h]
% \centering
% \resizebox{0.6\textwidth}{!}{%
% \begin{small}
% \begin{tabular}{|l|c|c|c|c|}
% \hline
%  & \textbf{Architecture} & \textbf{Layers} & \textbf{Dim} & \textbf{Params}\\
% \hline
% DPCNN~\cite{johnsonDeepPyramidConvolutional2017} & CNN & 4 & 256 & 2.9M\\
% \hline
% Stacked-LSTM~\cite{DeepLearningPython}  & RNN & 4 & 256 & 1.6M\\
% \hline
% BERT-Mini~\cite{devlinBERTPretrainingDeep2019}  & Transformer & 4 & 256 & 3.3M\\
% \hline
% % BERT-Base~\cite{devlinBERTPretrainingDeep2019}  & Transformer & 12 & 768 & 85M\\
% % \hline
% \end{tabular}
% \end{small}
% }
% \caption{Model architectures used to benchmark vectorizers. Parameters count excludes the parameters from the vectorizer and token embedding layers, since the number of parameters will vary depending on which vectorizer is used.}
% \label{tab:app:mod}
% \end{table}

\begin{table*}[t]
\resizebox{\textwidth}{!}{%
\begin{tabular}{|l|c|c|c|c|c|c|c|c|}
\hline
Name &
Train Size &
Test Size &
\# Classes &
\# Languages &
OOV Words &
Avg Sentence Length &
Avg Word Length\\
\hline
AG News~\cite{zhangCharacterlevelConvolutionalNetworks2016} & 120k & 7.6k & 4 & 1 & 9.9\% & 45 & 4.4\\
\hline
Yelp Reviews (Polarity) ~\cite{zhangCharacterlevelConvolutionalNetworks2016} & 560k & 38k & 2 & 1 & 17.4\% & 160 & 3.7\\
\hline
Multilingual Amazon Reviews (Polarity)~\cite{mcauleyHiddenFactorsHidden2013} & 960k & 24k & 2 & 6 & 35.2\% & 27 & 5.0\\
\hline
MASSIVE (Intent Classification) ~\cite{fitzgeraldMASSIVE1MExampleMultilingual2022} & 587k & 152k & 60 & 51 & 35.3\% & 6 & 5.1\\
\hline
\end{tabular}

}
\caption{List of datasets used for classification evaluation.}
\label{tab:datasets}
\end{table*}

In order to perform a comprehensive evaluation, we evaluate classification performance on four different datasets with drastically different dataset sizes, number of languages, classification tasks, and text lengths, as summarized in Table~\ref{tab:datasets}. For example, the MASSIVE~\cite{fitzgeraldMASSIVE1MExampleMultilingual2022} intent classification dataset has very short sentences but a lot of languages (51), whereas the Yelp Reviews dataset has significantly longer texts but is monolingual.

% We split examples on whitespace for all languages except for Chinese and Japanese for which we use Stanza tokenization~\cite{qiStanzaPythonNatural2020a}. For multilingual datasets, the training and test sets consist of all language splits.

We benchmark performance on three of the most common model architectures for text classification: Transformer (BERT-Mini~\cite{devlinBERTPretrainingDeep2019}), CNN (DPCNN~\cite{johnsonDeepPyramidConvolutional2017}
) and LSTM (Stacked-LSTM~\cite{DeepLearningPython}), details provided in Appendix~\ref{app:cla:mod}. To keep evaluation fair and consistent when comparing various vectorizers, we keep the models the same and only swap the vectorizers. All models are implemented in TensorFlow 2.11 and training is conducted on a Google Cloud VM using a single NVidia V100 GPU.

All models are trained with Adam optimizer with $\beta_1=0.9$, $\beta_2=0.999$ and a max learning rate of 5e-4, with the exception of fastText on multilingual datasets. We found that a max learning rate of 5e-4 was too high and led to divergence, so we used 1e-4 instead for fastText on MASSIVE and Multilingual Amazon Reviews. For BERT, we linearly warmup the learning rate for 5k steps. All models are trained for 100k steps with batch size 256 and cosine learning rate decay to 0. The test accuracy of each model averaged over 3 trials with different random seeds is reported in Table~\ref{tab:cla}.

\paragraph{Results}
% \begin{figure*}[t!]
%     \centering
%     \begin{subfigure}
%         \centering
%         \includegraphics[width=0.48\textwidth]{figures/classification_models.pdf}
%     \end{subfigure}
%     \hfill
%      \begin{subfigure}
%         \centering
%         \includegraphics[width=0.48\textwidth]{figures/classification_datasets.pdf}
%     \end{subfigure}
%     \caption{Classification performance.}
%     \label{fig:typo_detail}
% \end{figure*}

\begin{table*}[h]
\resizebox{\textwidth}{!}{%
\begin{tabular}{|c|c|c|c|c|c|c|c|}
\hline
\textbf{Dataset} & 
\textbf{Model} &
\textbf{Whitespace} &
\textbf{SentencePiece} &
\textbf{BPE} &
\textbf{fastText} &
\textbf{\rvr} &
\textbf{\rvb }
\\
\hline
\multirow{4}{*}{AG News} & RNN & 91.6\% & 91.3\% & 91.3\% & \textbf{92.7\%} & 91.3\% & \underline{92.6\%}  \\
 & CNN & \textbf{92.5\%} & \underline{92.4\%} & 90.7\% & 92.3\% & 88.2\% & 91.2\%  \\
 & BERT & 91.5\% & 91.5\% & 92.1\% & \underline{93.4\%} & 93.1\% & \textbf{93.5\%} \\
 & AVG & 91.9\% & 91.7\% & 91.4\% & \textbf{92.8\%} & 90.9\% & \underline{92.4\%}  \\
                         
\hline
\multirow{4}{*}{Yelp P.} & RNN & 93.9\% & 93.3\% & 93.4\% & \underline{94.7\%} & 94.1\% & \textbf{94.7\%}\\
 & CNN & \textbf{94.4\%} & 93.8\% & 93.0\% & \underline{93.9\%} & 92.5\% & 93.4\%  \\
 & BERT & 91.6\% & 91.3\% & 91.6\% & \textbf{93.5\%} & \underline{93.2\%} & 92.8\% \\
 & AVG & 93.3\% & 92.8\% & 92.7\% & \textbf{94.0\%} & 93.3\% & \underline{93.6\%} \\
 
\hline
\multirow{4}{*}{Multilingual Amazon P.} & RNN & \underline{93.2\%} & 90.7\% & 90.3\% & 87.0\% & 92.9\% & \textbf{93.5\%} \\
 & CNN & \textbf{93.2\%} & 90.8\% & 89.2\% & 84.4\% & 91.1\% & \underline{92.2\%} \\
 & BERT & 91.8\% & 87.3\% & 87.2\% & 86.6\% & \underline{92.7\%} & \textbf{92.6\%}  \\
 & AVG & \underline{92.8\%} & 89.6\% & 88.9\% & 86.0\% & 92.2\% & \textbf{92.8\%} \\
                         
\hline
\multirow{4}{*}{MASSIVE} & RNN & 70.3\% & 69.6\% & 68.6\% & 13.5\% & \underline{71.0}\% & \textbf{73.8\%}\\
 & CNN & \textbf{70.4\%} & \underline{69.2\%} & 59.2\% & 12.6\% & 59.8\% & 66.7\%  \\
 & BERT & 70.6\% & 67.8\% & 66.2\% & 23.9\% & \underline{78.1}\% & \textbf{78.6}\%  \\
 & AVG & \underline{70.4\%} & 68.9\% & 64.7\% & 16.7\% & 69.6\% & \textbf{73.0\%}\\
                         
\hline
\multirow{4}{*}{Average} & RNN & \underline{87.3}\% & 86.2\% & 85.9\% & 72.0\% & 87.3\% & \textbf{88.7\%} \\
 & CNN & \textbf{87.6\%} & \underline{86.5\%} & 83.0\% & 70.8\% & 82.9\% & 85.9\% \\
 & BERT & 86.4\% & 84.5\% & 84.3\% & 74.4\% & \underline{89.3}\% & \textbf{89.4\%} \\
 & AVG & \underline{87.1\%} & 85.8\% & 84.4\% & 72.4\% & 86.5\% & \textbf{88.0\%}  \\
\hline
\end{tabular}
}
\caption{Detailed classification results comparing different vectorizers when used to train models from scratch. \textbf{Bold} indicates best results, \underline{underline} indicates second best.}
\label{tab:cla}
\end{table*}

\begin{figure*}[h]
    \centering{
    \includegraphics[width=0.9\columnwidth]{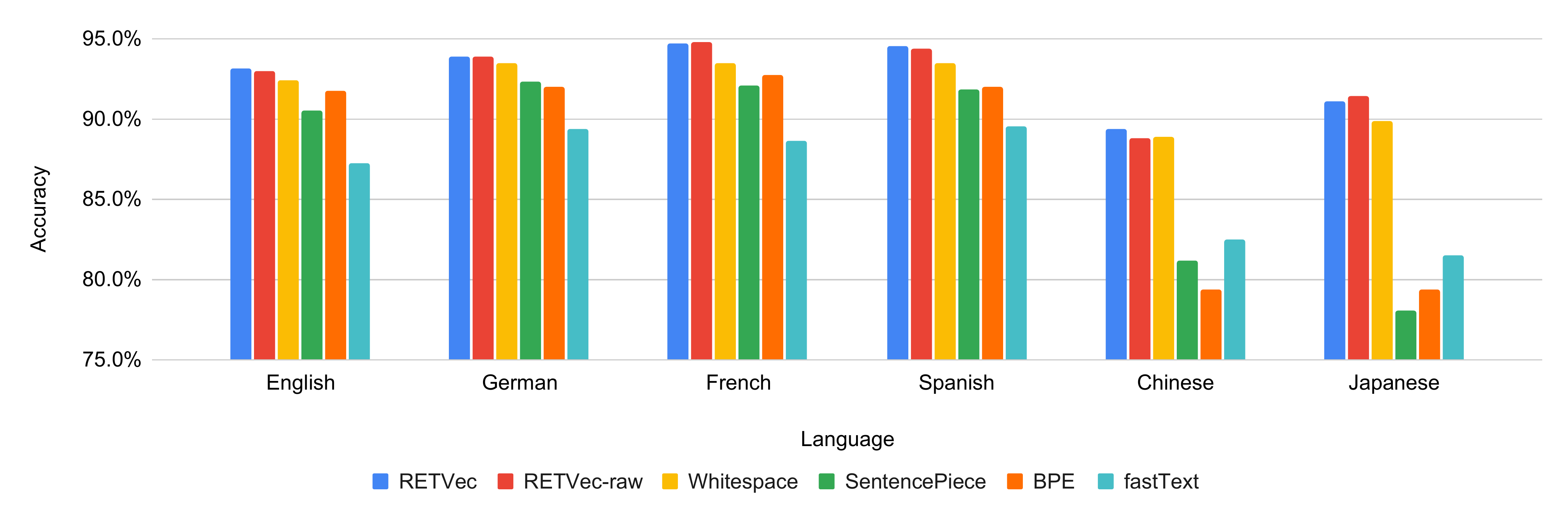}
    \caption{Classification performance on Multilingual Amazon Reviews broken down by language.}
    \label{fig:amazon}
    }
\end{figure*}

Overall, averaging over all datasets and model architectures, \rvb is the best performing vectorizer by some margin (0.9\%). We observe that \rv performs best when paired with Transformer and RNN architectures, but lags slightly behind for the CNN. Compared to \rvr, \rv's pre-trained model offers a significant performance improvement across all datasets and model architectures. 

fastText performs the best on average for the English-only datasets, but does not perform well on the two multilingual datasets (especially MASSIVE), which explains its overall poor performance. Given that we are using the same code and models for all datasets and the fact that fastText results on English-only datasets match the ones reported in the original paper~\cite{joulinBagTricksEfficient2017}, we believe that fastText's poor performance on multilingual datasets is due to the fact that fastText vectors from different languages may not be compatible with each other when used as inputs to a multilingual model. However, we didn’t find any mention of such an issue in the literature.

Delving deeper into multilingual capabilities, as plotted in Figure~\ref{fig:amazon}, we observe that \rv consistently outperforms other vectorizers regardless of the language. We also observe that \rv has a wider lead on non-Latin languages, which we attribute at least partially to our novel character encoder, given that \rv and \rvr achieve similar performance on them.

\section{Evaluation: Typo Resilience}
\label{sec:typ}
\label{sec:res}

\begin{figure}[h]
    \centering

    \includegraphics[width=0.48\textwidth]{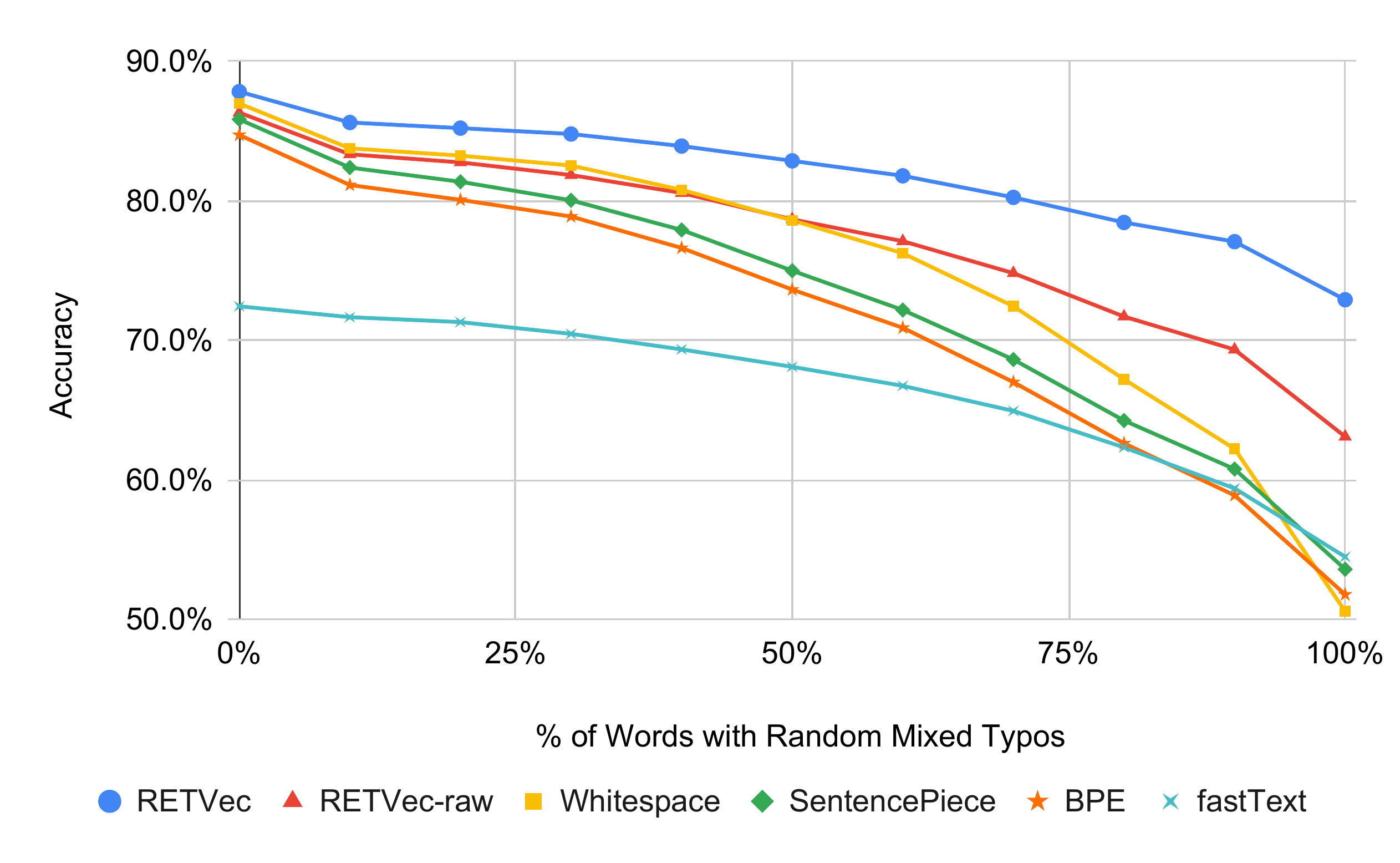}
    \hfill
    \includegraphics[width=0.48\textwidth]{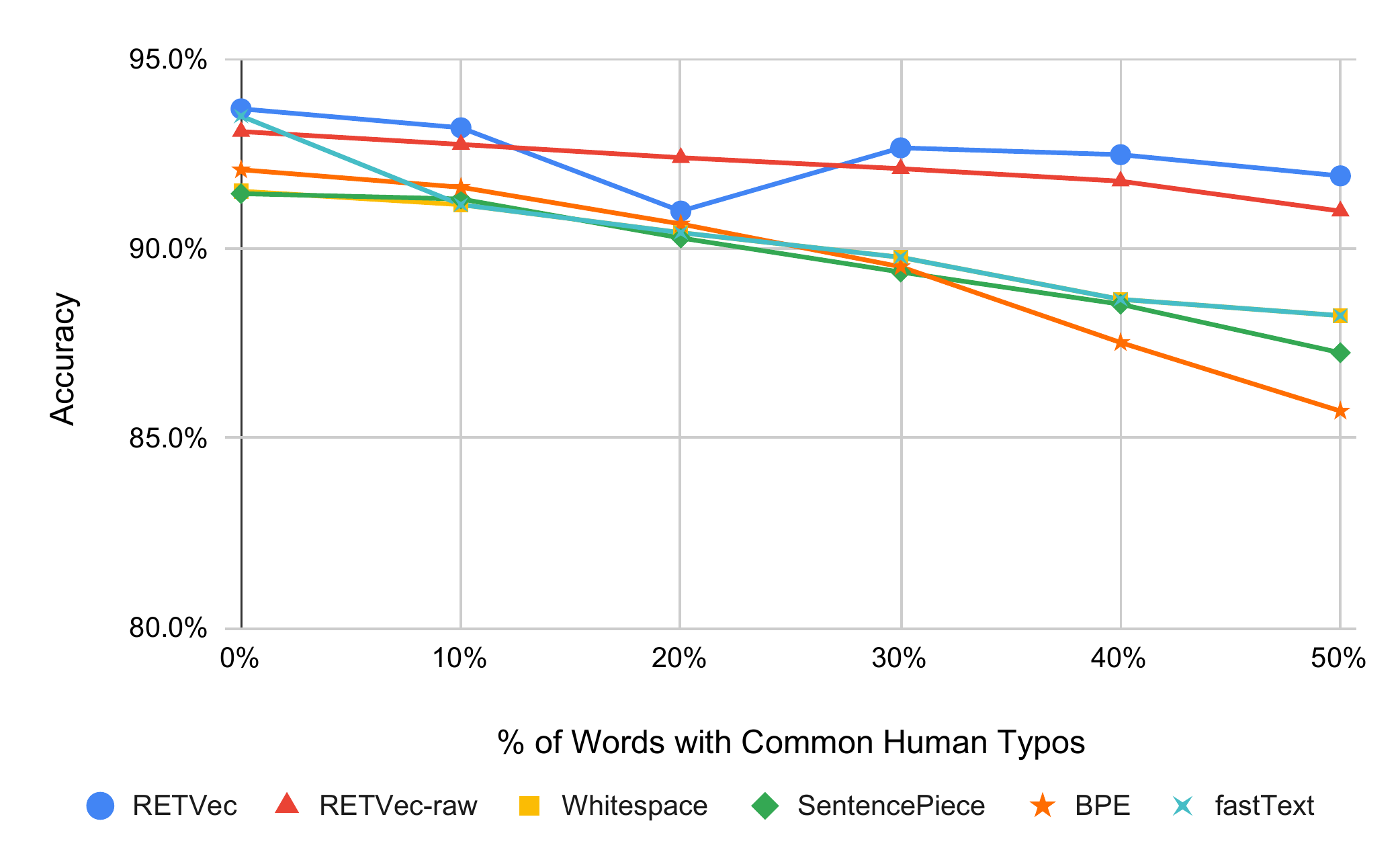}

    \caption{Comparison of various tokenizers' resilience against mixed random typos (left) and common human typos (right) when training classification models from scratch.}
    \label{fig:typo_cla}
\end{figure}

% \begin{table*}[h]
% \centering
% \resizebox{0.6\textwidth}{!}{%
% \begin{tabular}{|l|r|r|r|r|r|r|}
% \hline
% \textbf{Vectorizer} & \textbf{0\%}    & \textbf{10\%}   & \textbf{20\%}   & \textbf{30\%}   & \textbf{40\%}   & \textbf{50\%}   \\ \hline
% Whitespace          & 91.5\%          & 91.2\%          & 90.4\%          & 89.8\%          & 88.7\%          & 88.2\%          \\ \hline
% SentencePiece       & 91.5\%          & 91.3\%          & 90.3\%          & 89.4\%          & 88.5\%          & 87.3\%          \\ \hline
% BPE                 & 92.1\%          & 91.6\%          & 90.7\%          & 89.5\%          & 87.5\%          & 85.7\%          \\ \hline
% fastText            & \underline{ 93.5\%}    & 91.2\%          & 90.4\%          & 89.8\%          & 88.7\%          & 88.2\%          \\ \hline
% RETVec-raw          & 93.1\%          & \underline{ 92.8\%}    & \underline{ 92.4\%}    & \underline{ 92.1\%}    & \underline{ 91.8\%}    & \underline{ 91.0\%}    \\ \hline
% RETVec              & \textbf{93.7\%} & \textbf{93.2\%} & \textbf{91.0\%} & \textbf{92.7\%} & \textbf{92.5\%} & \textbf{91.9\%} \\ \hline
% \end{tabular}
% }
% \caption{Classification results on human-like typos using Neuspell. \textbf{Bold} indicates best results, \underline{underline} indicates second best.}
% \label{tab:human-typo}
% \end{table*}

% \begin{figure}[t!]
%     \centering
%     \includegraphics[width=0.49\textwidth]{figures/typo_human.pdf}
%     \caption{Fixme move as table}
%     % \label{fig:typo_human}
%     \label{fig:typo_detail}
% \end{figure}

\begin{figure}[h]
    \centering
    \begin{subfigure}
        \centering
        \includegraphics[width=0.48\textwidth]{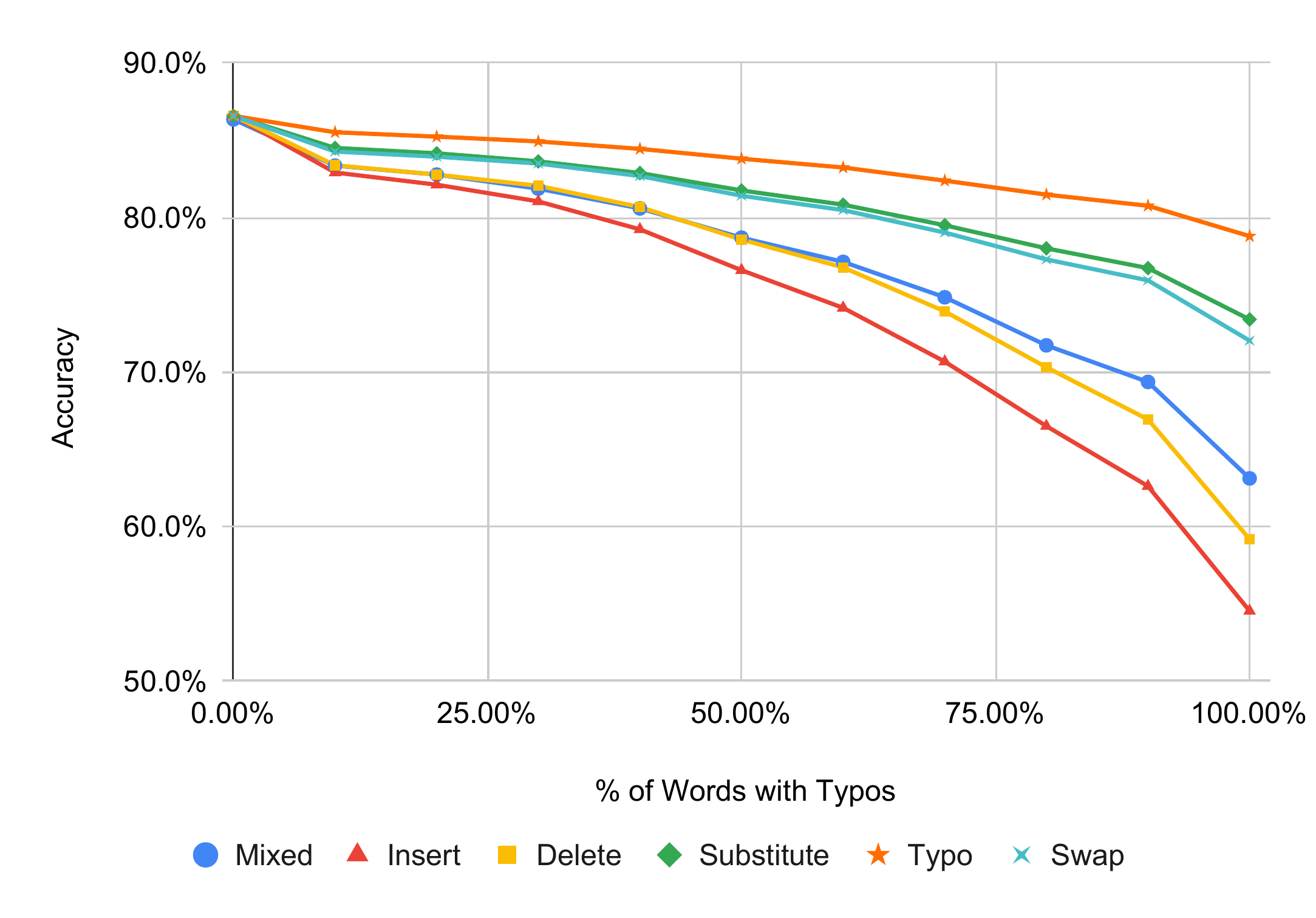}
    \end{subfigure}
    \hfill
     \begin{subfigure}
        \centering
        \includegraphics[width=0.48\textwidth]{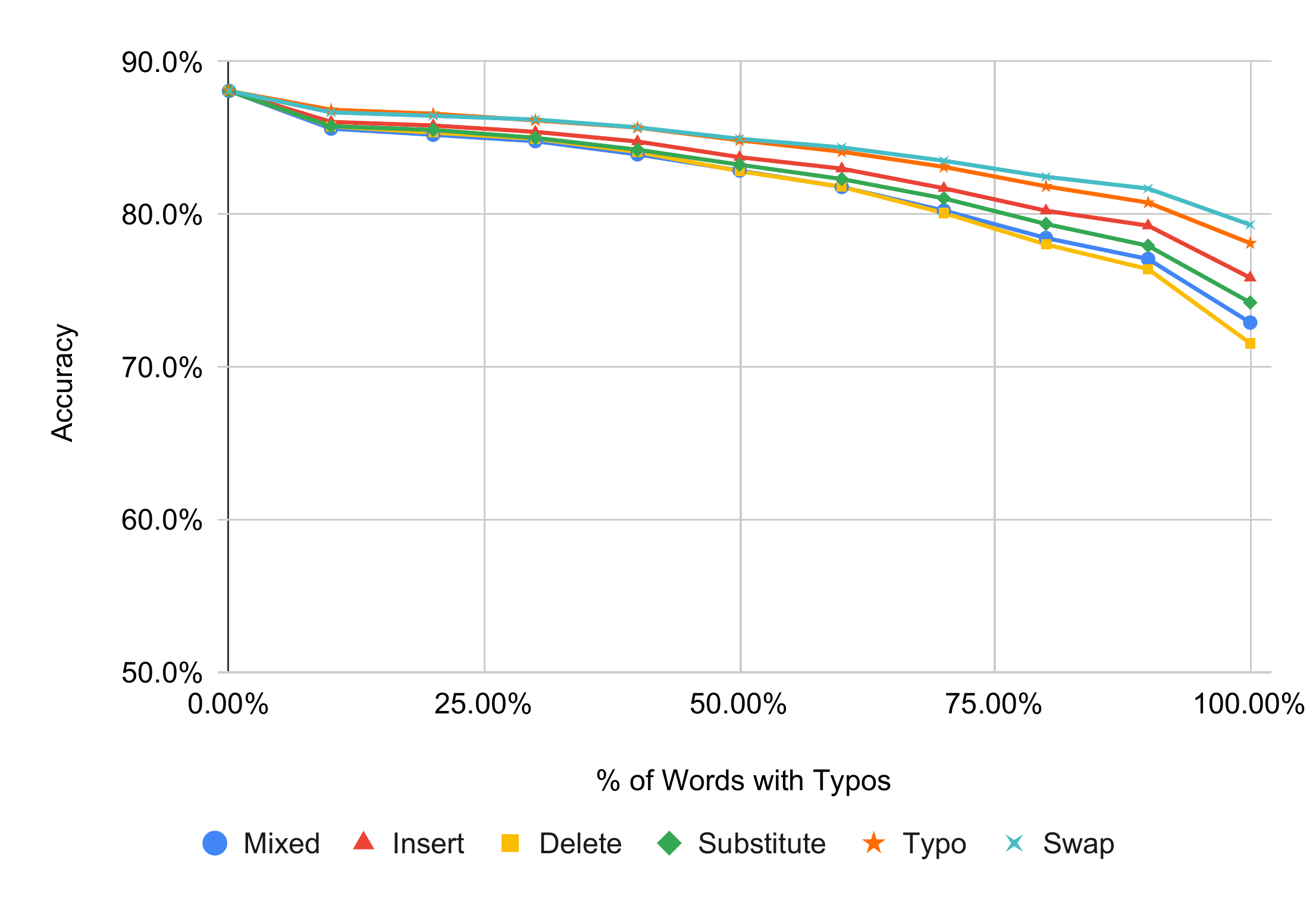}
    \end{subfigure}

    \caption{Comparison of \rv resilience against various types of typos. \rvr on the left, \rvb on the right.}
    \label{fig:typo_detail}
\end{figure}

% \begin{table*}[t]
% \resizebox{\textwidth}{!}{%
% \begin{tabular}{|l|r|r|r|r|r|r|r|r|r|r|r|}
% \hline
% Typo &0\% &10\% &20\% &30\% &40\% &50\% &60\% &70\% &80\% &90\% &100\% \\\hline
% SentencePiece &88.3\% &85.4\% &84.4\% &83.1\% &80.9\% &77.8\% &74.5\% &70.3\% &64.8\% &59.4\% &49.9\% \\\hline
% BPE &87.4\% &84.4\% &83.4\% &82.1\% &79.9\% &76.8\% &73.7\% &69.3\% &63.8\% &58.3\% &48.6\% \\\hline
% fastText &77.2\% &76.6\% &76.2\% &75.4\% &74.3\% &73.0\% &71.5\% &69.4\% &66.0\% &61.8\% &53.7\% \\\hline
% Whitespace &89.3\% &86.7\% &86.1\% &85.3\% &83.6\% &81.3\% &78.7\% &74.3\% &68.1\% &61.2\% &46.4\% \\\hline
% \rvr &88.6\% &86.1\% &85.5\% &84.7\% &83.4\% &81.6\% &79.9\% &77.5\% &74.4\% &71.6\% &65.2\% \\\hline
% \rvb &\uline{89.5\%} &\uline{87.9\%} &\uline{87.6\%} &\uline{87.3\%} &\uline{86.7\%} &\uline{85.9\%} &\uline{85.3\%} &\uline{84.2\%} &\uline{83.1\%} &\uline{82.3\%} &\uline{79.8\%} \\\hline
% \rvl &\textbf{89.5\%} &\textbf{87.9\%} &\textbf{87.7\%} &\textbf{87.4\%} &\textbf{86.7\%} &\textbf{86.1\%} &\textbf{85.4\%} &\textbf{84.3\%} &\textbf{83.1\%} &\textbf{82.3\%} &\textbf{79.9\%} \\\hline
% \end{tabular}
% }
% \caption{\fix{} Average classification accuracy for the models trained in Section~\ref{sec:cla} as the fraction of  the  words corrupted in each test sample increases. Bold indicates best results, underlined values are the 2nd best.}
% \label{tab:typo-res}
% \end{table*}

In this section, we evaluate the resilience of the various vectorizers against both random typos and common human typos.

\paragraph{Setup}

We rely on two types of typo injection methods to evaluate the tokenizers' typo resilience: {\tt random typos} and {\tt common human typos}.
For random typos, evaluation is performed on every model and dataset used in Section~\ref{sec:cla} and average accuracy on the typo-augmented test splits is recorded. For common human typos, because we only have them for English, we take the average accuracy across all three models on only AG News.
We evaluate how each tokenizer's resilience degrades as the number of typos increases by gradually increasing the percentage of words with typos in the test set by increments of 10\%.

Random typos are created by using a combination of insertion, deletion, substitution, neighboring swap, and keyboard-based typos. Each typo is applied with a block size of 1 or 2 characters. Words are selected at random and each word can only have one typo. To create human-like typos, we use the word replacement {\tt noiser} from the Neuspell~\cite{jayanthiNeuSpellNeuralSpelling2020} package. The noiser uses 109k common misspellings of 17k popular English words. On the AG News test set, it can only inject typos into ~55\% of the words, so we limit the evaluation to 50\% words with common human typos.

\paragraph{Results}

As reported in Figure~\ref{fig:typo_cla}, the performance of \rv and \rvr decreases significantly slower than the other vectorizers as the amount of typos increase. \rv's pre-trained model increases \rv resilience by up to 15\% compared to \rvr, demonstrating the effectiveness of \rv's pre-training technique of using pair-wise metric learning to create syntactically robust word embeddings. SentencePiece and BPE perform about the same, with a gradual decline which keeps them competitive with \rvr until about 60\% random typo rate. We note that fastText and Whitespace provide more typo resilience than SentencePiece and BPE, which is expected because it has been previously noted that word-level vectorizers are typically more robust than character-level or subword-level vectorizers~\cite{pruthiCombatingAdversarialMisspellings2019}.  We see similar trends on the common human typo results, with \rvr and \rvb being noticeably more resilient than baseline vectorizers when the typo rate is >30\%.

Delving deeper into \rv's typo resilience capabilities, as reported in Figure ~\ref{fig:typo_detail}, character deletion is the hardest form of typo for \rv to deal with. Without the pre-trained model, \rvr also struggles with character insertion and \rv's model also greatly helps nullify the impact of character swapping.
\section{Evaluation: Adversarial Attack Resilience}
\label{sec:adv}
\begin{table*}[t]
\centering
    \resizebox{0.6\textwidth}{!}{%
    \footnotesize
\begin{tabular}{|l|rrr|}
\hline
\textbf{Tokenizer} & \multicolumn{1}{c|}{\textbf{Original Acc}} & \multicolumn{1}{c|}{\textbf{Acc under Atk}} & \multicolumn{1}{c|}{\textbf{Atk Success \%}}\\ \hline
Whitespace & \multicolumn{1}{r|}{89.4\%} & \multicolumn{1}{r|}{32.3\%} & 63.9\%  \\ \hline
SentencePiece & \multicolumn{1}{r|}{89.8\%} & \multicolumn{1}{r|}{25.8\%} & 71.3\%  \\ \hline
BPE & \multicolumn{1}{r|}{90.8\%} & \multicolumn{1}{r|}{29.2\%} & 67.8\%  \\ \hline
fastText & \multicolumn{1}{r|}{92.6\%} & \multicolumn{1}{r|}{40.5\%} & 56.3\% \\ \hline
RETVec-raw & \multicolumn{1}{r|}{\underline{93.0\%}} & \multicolumn{1}{r|}{\underline{50.6\%}} & \underline{45.6\%} \\ \hline
RETVec & \multicolumn{1}{r|}{\textbf{93.7\%}} & \multicolumn{1}{r|}{\textbf{51.7\%}} & \textbf{44.8\%} \\ \hline
\end{tabular}
%     \begin{tabular}{|l|r|r|r|r|r|r|r|r|r|r|r|r|}
%     \hline
%     \textbf{    Vectorizer} &\textbf{SentencePiece}&\textbf{BPE} &\textbf{fastText}& \textbf{Whitespace} &\textbf{\rvr} &\textbf{\rvb} &\textbf{\rvl} \\\hline
%     Base accuracy &89.9\% &90.8\% &92.6\% &89.4\% &92.2\% &93.9\% &92.9\% \\\hline
%     Acc under Attack &23.1\% &29.0\% &40.5\% &32.4\% &51.2\% &58.8\% &56.0\% \\\hline
%     Accuracy reduction &-66.8\% &-61.8\% &-52.1\% &-57.0\% &-41.0\% &\textbf{-35.1\%} &\uline{-36.9\%} \\\hline
%     Attack success rate &74.3\% &68.1\% &56.3\% &63.7\% &44.5\% &\textbf{37.4\%} &\uline{39.8\%} \\\hline
% \end{tabular}
}
\caption{Adversarial resilience evaluation for BERT-Mini trained on AG News with different vectorizers. Adversarial attack results are averaged across three types of character-level adversarial attacks: TextBugger ~\cite{liTextBuggerGeneratingAdversarial2019}, DeepWordBug ~\cite{gaoBlackboxGenerationAdversarial2018}, and Pruthi ~\cite{pruthiCombatingAdversarialMisspellings2019}. \textbf{Bold} indicates best results, \underline{underline} indicates second best.}
\label{tab:unp:adv}
\end{table*}

\paragraph{Setup}
We use the TextAttack framework~\cite{morrisTextAttackFrameworkAdversarial2020} to evaluate the resilience of \rv and other vectorizers against character-level adversarial attacks, specifically:  TextBugger~\cite{liTextBuggerGeneratingAdversarial2019}, DeepWordBug~\cite{gaoBlackboxGenerationAdversarial2018}, and Pruthi~\cite{pruthiCombatingAdversarialMisspellings2019}. For each model and vectorizer, the attacks are carried out on the same 1000 examples drawn randomly from the AG News test dataset.

%The table reports the following metrics:
%\textbf{Base accuracy}: The classification accuracy on the original, unaugmented examples.
%\textbf{Acc under attack}: The classification accuracy on the augmented adversarial examples.
%\textbf{Accuracy reduction}: The difference between the base accuracy and the accuracy under attack.
%\textbf{Attack success rate}: The percentage of augmented examples that resulted in a successful change from %TP to FP.

\paragraph{Results}

% In Table~\ref{tab:unp:adv}, we report the average classification results over the 1000 adversarial examples, with detailed results in the appendix (table~\ref{tab:app:adv:ag:bert2}).

Table~\ref{tab:unp:adv} demonstrates that \rv models are significantly more resilient to adversarial text attacks than other embedding schemes. In particular, the average attack success rate against \rvr and \rvb models are 45.6\% and 44.8\% respectively, which is significantly lower than for SentencePiece (71.3\%) and BPE (67.8\%). The best performing, non-\rv, vectorizer is fastText (56.3\% attack success rate), which makes sense because fastText handles OOV tokens by averaging the vector representations of the token's n-grams. Detailed results broken down by adversarial attack algorithm are reported in Appendix~\ref{app:adv}.

% Surprisingly, the attack success rate is higher against \rvl (39.8\%) than it is for \rvb (37.4\%). As both models are trained in the same manner, this seems to imply that the transformer architecture is less resilient to text perturbations. This is unexpected as the dense architecture of \rvb has a higher final training loss compared to \rvl (see table~\ref{ref:retvec-overview}). We aren’t aware of a good explanation for this phenomenon.

\section{Evaluation: Pre-training BERT}
\label{sec:large}
In this section, we evaluate \rv's performance when used to pre-train transformers. We evaluate both \rv's competitiveness on the GLUE benchmark~\cite{wang2019glue} and typo resilience.

\paragraph{Setup}
We use BERT-Base~\cite{devlinBERTPretrainingDeep2019} for all experiments. Given that \rv outputs word-level embeddings and is not restricted by a vocabulary size, we cannot directly use the standard BERT masked-language modeling (MLM) task and predict token ids using a softmax layer. Instead, we use an approximate MLM task for \rv-based models by only masking and predicting input tokens that are in the top 100k words. Other than this, we follow the standard MLM procedure by selecting 15\% of tokens at random and replacing a selected token with (1) the [MASK] token with 80\% chance (2) a random token with 10\% chance (3) the same token unchanged with 10\% chance. For the baseline, we use SentencePiece with a vocabulary size of 32k and the standard MLM task described in ~\cite{devlinBERTPretrainingDeep2019}.

For each model, we pre-train for 1M steps with batch size 64 on the English C4 dataset~\cite{xueByT5TokenfreeFuture2022}. We use Adam with a max learning rate of 5e-5, $\beta_1=0.9$, $\beta_2=0.999$, and $0.01$ L2 weight decay. We warmup for 10000 steps and linearly decay the learning rate. We fine-tune all models for 20 epochs on GLUE using 3 different random seeds and report the best result for each dataset. We pre-train and fine-tune each model using 8 NVidia V100s. Detailed pre-training and fine-tuning hyperparameters can be found in Appendix~\ref{app:pretrain_bert}.

To benchmark typo resilience of \rv-based pre-trained BERT models, we fine-tune on AG News to produce comparable results to Section~\ref{sec:res}, and follow the same evaluation methodology against random and common human typos.

\paragraph{Results}

% \begin{table*}[t]
% \resizebox{\textwidth}{!}{%
% \label{tab:cla}
% \end{table*}

\begin{table*}[h]
\centering
  \scalebox{0.9}{%
\begin{tabular}{|l|rrrrrrrrr|}
\hline
      \textbf{ Vectorizer}  & \multicolumn{1}{l}{\textbf{MNLI}} & \multicolumn{1}{l}{\textbf{QNLI}} & \multicolumn{1}{l}{\textbf{QQP}} & \multicolumn{1}{l}{\textbf{RTE}} & \multicolumn{1}{l}{\textbf{SST-2}} & \multicolumn{1}{l}{\textbf{MRPC}} & \multicolumn{1}{l}{\textbf{CoLA}} & \multicolumn{1}{l}{\textbf{STS-B}} & \multicolumn{1}{l|}{\textbf{Avg}} \\ \hline
SentencePiece               & 80.6                     & 88.7                     & 90.1                    & \textbf{67.2}           & \underline{91.1}        & 85.8                     & \textbf{50.7}            & \textbf{82.5}             & \textbf{79.6}                 \\ \hline 
\rvr                        & \textbf{82.5}            & \textbf{89.6}            & \textbf{90.5}           & 65.3                    & \textbf{91.5}             & \underline{87.2}       & \underline{49.1}           & 79.3                      & \underline{79.4}                          \\ \hline
\rvb                        & \underline{81.4}   & \underline{89.2}          & \underline{90.4}       & \underline{65.7}         & 90.8                      & \textbf{87.3}         & 47.9                     & \underline{79.8}                          & 79.1                          \\ \hline
\end{tabular}
}
\caption{Results on GLUE dev sets. We report matched accuracy for MNLI, Matthews correlation for CoLA, Pearson correlation for STS-B, F1 score for QQP, and accuracy for all other tasks. \textbf{Bold} indicates best results, \underline{underline} indicates second best.}
\label{tab:large:glue}
\end{table*}

\begin{figure}[h]
    \centering
    \begin{subfigure}
        \centering
        \includegraphics[width=0.48\textwidth]{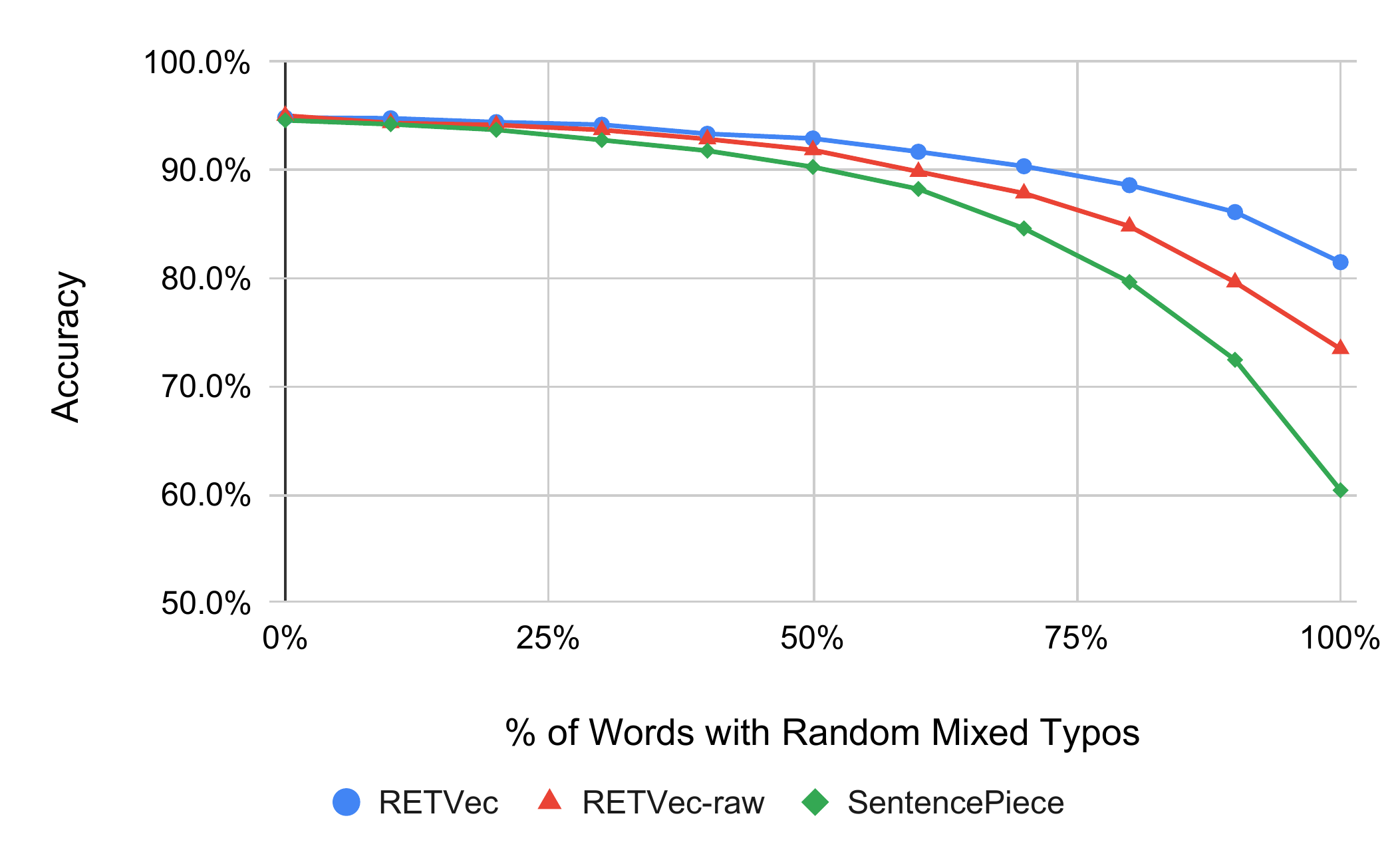}
    \end{subfigure}
    \hfill
     \begin{subfigure}
        \centering
        \includegraphics[width=0.48\textwidth]{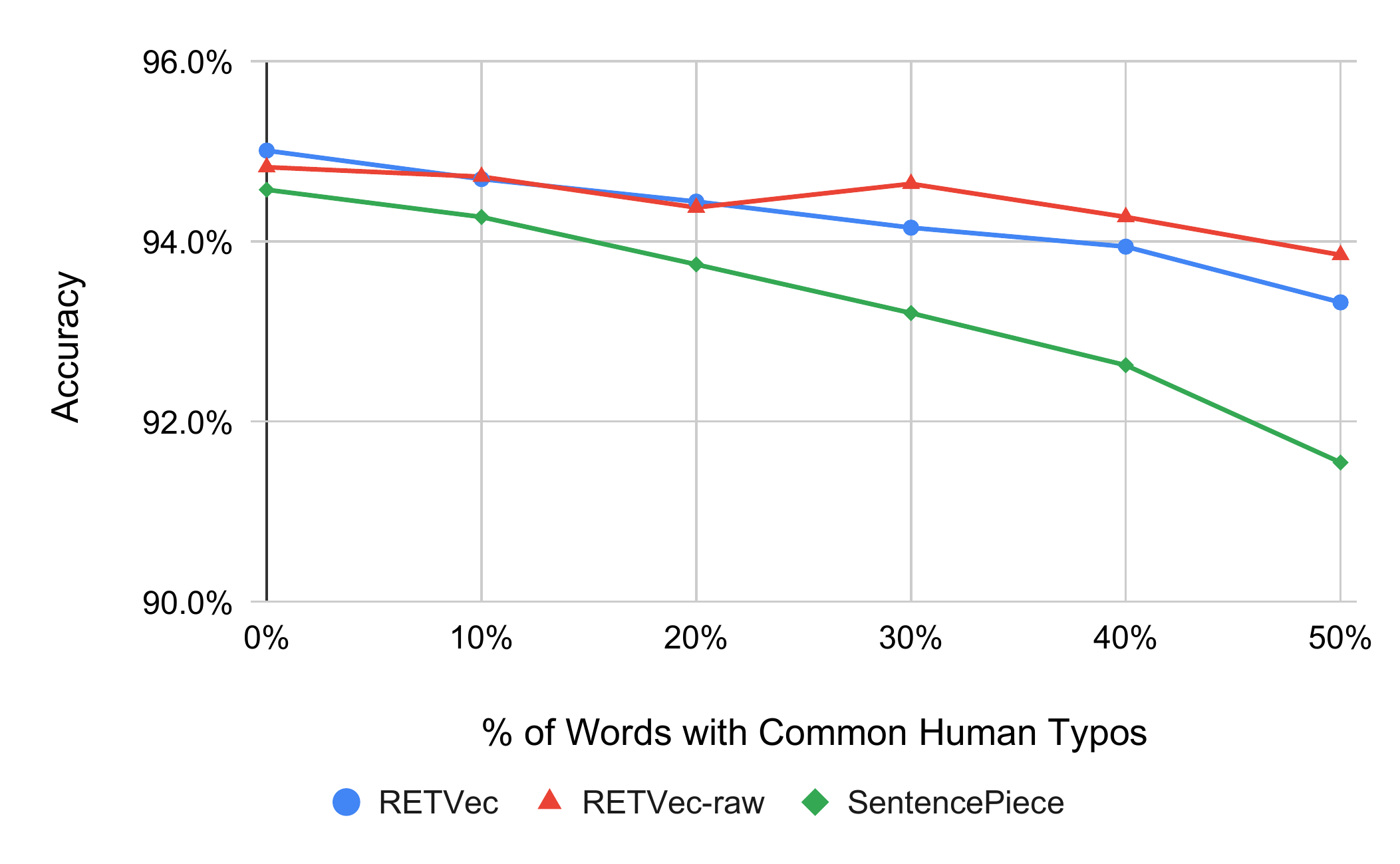}
    \end{subfigure}

    \caption{Comparison of typo resilience against mixed random typos (left) and common human typos (right) when fine-tuning a pre-trained BERT-Base model on AG News dataset for \rv and SentencePiece.}
    \label{fig:bert_typo}
\end{figure}

Overall, as reported in Table~\ref{tab:large:glue}, BERT pre-trained with \rv and \rvr using MLM on 100k words achieves competitive results compared to SentencePiece. Additionally, using \rv reduces the pre-training time of BERT by around 10\% in our experiment, when training for an equivalent number of steps. Surprisingly, \rvr is more competitive than \rv, which we attribute to the fact that BERT-Base has more than enough capacity to learn a competitive representation without \rv's word embedding model. However, this increased competitiveness for \rvr comes at the cost of decreased typo resilience, as reported in Figure~\ref{fig:bert_typo}, although the difference in typo resilience between pre-trained BERT models using \rvr versus \rvb is only significant when the percentage of injected typos exceeds 50\% of the words. As such, it seems that using \rvr when pre-training large models is the best choice. How to optimally do so is left for future work.
\section{Evaluation: In the Wild}
\label{sec:wil}
As a final evaluation, to ensure \rv will work in practice against adversarial content, we trained a version of a transformer model we use as part of our spam email filtering system at Google. As with previous experiments, we kept everything constant besides swapping SentencePiece with \rv. Benchmarking on our internal production evaluation dataset, we found that \rv improved the recall at 0.80\% false positive rate by $\sim$0.51\%, helping detect 38\% of the spam emails that would have been missed by the baseline model while reducing model serving latency by 30\%. This result gives us confidence that \rv is competitive for real-world classification tasks, especially in the face of adversarial content.
\section{Ablation Study}
\label{sec:abl}

In this section, we report how key hyperparameters variations affect \rv's performance.
An unforeseen challenge we discovered while developing \rv models is that a lower pre-training loss didn’t directly translate into meaningful performance on our benchmarks. To overcome this challenge, we had to include an extra validation step that trains and evaluates models on the Multilingual Amazon Reviews dataset~\cite{mcauleyHiddenFactorsHidden2013} using the methodology described in Section~\ref{sec:cla} to measure downstream improvement. We choose the Amazon dataset due to its multilingual nature and difficulty. Detailed results on classification performance by model type, pre-training loss, and model size for the ablation studies can be found in Appendix~\ref{app:abl}, along with additional experiments on less impactful hyperparameters.

\begin{table}[h]
\resizebox{\textwidth}{!}{%
\footnotesize
\begin{tabular}{|c|c|c|}

\hline
\multicolumn{1}{|l|}{\textbf{Architecture}} & \multicolumn{1}{c|}{\textbf{Loss}} & \multicolumn{1}{c|}{\textbf{Accuracy}} \\ \hline
\textbf{MLP} & 0.0248 & 92.89\% \\ \hline
BERT + MLP & 0.0129 & 92.54\% \\ \hline
T5 + MLP & 0.0120 & 92.72\% \\ \hline
GAU + MLP & 0.0133 & 92.82\% \\ \hline
LSTM + MLP & 0.0179 & 92.61\% \\ \hline
CNN + MLP & 0.0214 & 92.72\% \\ \hline
% \begin{tabular}{|c|c|c|}
% \hline
% \multicolumn{1}{|l|}{\textbf{Activation}} & \multicolumn{1}{|c|}{\textbf{Loss}} & \multicolumn{1}{|c|}{\textbf{Acc}} \\ \hline
% \textbf{Tanh}                                              & 0.0248                                       & 92.9\%                                 \\ \hline 
% Softsign                                           & 0.0259                                       & 92.7\%                                          \\ \hline 
% ReLU1                                            & 0.0233                                       & 92.6\%                                          \\ \hline
% Sigmoid                                            & 0.0246                                       & 92.6\%                                          \\ \hline
\end{tabular}

% \caption{RETVec embedding activation ablation study. \textbf{Bold} denotes the hyperparameter selected for the final \rvb model.}
% \label{tab:abl:act}
% \end{table}
\quad
% \begin{table}[t]
% \resizebox{\columnwidth}{!}{%
% \footnotesize
\begin{tabular}{|c|c|c|}
\hline
\multicolumn{1}{|l|}{\textbf{Emb Dim}} & \multicolumn{1}{|c|}{\textbf{Loss}} & \multicolumn{1}{|c|}{\textbf{Acc}} \\ \hline

%64 & 0.0284 & 92.4\%                                          \\ 
%\hline
100                                         & 0.0267                                        & 92.6\%                                          \\ \hline 
128                                         & 0.0260                                        & 92.7\%                                          \\ \hline 
200                                         & 0.0253                                        & 92.7\%                                          \\ \hline 
\textbf{256}                                        & 0.0248                                        & 92.9\%                                 \\ \hline 
300                                         & 0.0245                                        & 92.8\%                                          \\ \hline 
384                                         & 0.0237                                        & 92.8\%                                          \\ \hline
% 512                                         & 0.0225                                        & 92.9\%  \\ \hline                                        
\end{tabular}

\quad

\begin{tabular}{|r|r|r|}
\hline
\multicolumn{1}{|l|}{\textbf{Word Len}} & \multicolumn{1}{c|}{\textbf{Loss}} & \multicolumn{1}{c|}{\textbf{Acc}} \\ \hline
% 8  & 0.04281 & 92.85\% \\ \hline
% 10 & 0.03263 & 92.83\% \\ \hline
12 & 0.0267 & 92.8\% \\ \hline
14 & 0.0259 & 92.8\% \\ \hline
\textbf{16} & 0.0248 & 92.9\% \\ \hline
20 & 0.0242 & 92.7\% \\ \hline
24 & 0.0234 & 92.9\% \\ \hline
% 28 & 0.0246  & 92.78\% \\ \hline
32 & 0.0235 & 92.8\% \\ \hline
\end{tabular}
}
\space
\caption{Ablation studies for \rv architecture (left), word embedding dimension (middle), and word length (right). \textbf{Bold} denotes the hyperparameter selected for the final \rvb model.}
\label{tab:abl}
\end{table}

\paragraph{Model architecture} Despite considerable efforts, we couldn't find a more complex architecture that outperformed the simple MLP architecture (Table~\ref{tab:abl}).
The best transformer-based architecture which uses 2 encoder blocks of dimension 128 resulted in significantly lower loss regardless of the transformer block used (BERT~\cite{devlinBERTPretrainingDeep2019}, GAU~\cite{huaTransformerQualityLinear2022}, T5~\cite{xueByT5TokenfreeFuture2022}). However, as alluded to earlier, the significantly lower loss did not translate to better performance on the benchmarks. We hypothesize that the fitting the text manifold is an easy enough task to solve with any reasonable architecture.

\paragraph{Embedding Size} 

We train \rvb with different embedding dimensions and report the results in Table~\ref{tab:abl}. Classification benchmarks show little difference in performance for embedding dimensions between 
128 and 512, with 256 embedding dimension having the slightest edge.

\paragraph{Max Input Character length}

To select the optimal max word length for \rv, we started by looking the fastText dataset's word length distribution. More than 95\% of the words are less than 16 characters long and the median word length is 7.9 characters (Appendix~\ref{app:fasttext}). Using more than 16 characters didn't led to any accuracy improvements despite those longer inputs exhibiting a lower loss, as reported in Table~\ref{tab:abl}.

\section{Discussion and Future Work}
\label{sec:con}

In this paper, we presented \rv, a new multilingual text vectorizer that combines a novel character encoding with an small model to project words into a compact, 256-dimensional embedding. Through extensive evaluations, we demonstrated that models trained using \rv achieve state-of-the-art classification performance and provide a 10-15\% increase in resilience against typos and adversarial attacks compared to popular text vectorizers and word embeddings.

The key outstanding question is how to best utilize \rv for generative tasks and in large language models (LLMs), which could enable us to train LLMs with stronger multilingual capabilities, improved adversarial robustness, and reduced model size and computational costs. In particular, for “small” LLMs (less than or around 1 billion total parameters), the vocabulary embedding layer can be more than 20\% of the total parameters~\cite{biderman2023pythia}, which would be eliminated when using \rv.

We discovered that the main difficulty when training generative models using \rv is that \rv's 256-float embedding cannot be directly converted into a softmax output, unlike other tokenizers which represent text as integer token IDs. In order to effectively use \rv in decoder-only models, we expect that a new training procedure for \rv-based models which is compatible with text generation tasks needs to be invented, e.g. a pre-training procedure which does not rely on directly predicting the next token. We briefly experimented with ideas including decoding the \rv representation character-by-character or using a VQ-VAE~\cite{oord2018neural} model to quantize the output, but the results so far were inconclusive. We hope to explore this area and address this limitation in future work. Other potential directions of future work include using \rv as a word embedding in various applications in place of GloVe~\cite{penningtonGloVeGlobalVectors2014} and word2vec~\cite{mikolovEfficientEstimationWord2013}, and using the \rv character encoder and training procedure to train text similarity models.

\bibliography{retvec}

%%%%%%%%%%%%%%%%%%%%%%%%%%%%%%%%%%%%%%%%%%%%%%%%%%%%%%%%%%%%%%%%%%%%%%%%%%%%%%%
%%%%%%%%%%%%%%%%%%%%%%%%%%%%%%%%%%%%%%%%%%%%%%%%%%%%%%%%%%%%%%%%%%%%%%%%%%%%%%%
% APPENDIX
%%%%%%%%%%%%%%%%%%%%%%%%%%%%%%%%%%%%%%%%%%%%%%%%%%%%%%%%%%%%%%%%%%%%%%%%%%%%%%%
%%%%%%%%%%%%%%%%%%%%%%%%%%%%%%%%%%%%%%%%%%%%%%%%%%%%%%%%%%%%%%%%%%%%%%%%%%%%%%%
\newpage
\appendix
\onecolumn
\section*{Appendix}
\vspace{10pt}

\section{\rv Model Details}
\label{app:arch}

Table~\ref{tab:app:rvb} details the hyperparameter settings for the \rv model architecture, as described in Section~\ref{sec:arc}.

\begin{table}[h]
    \centering
    \begin{tabular}{|l|r|}
    \hline
\footnotesize
\textbf{Hyperparameter} &\textbf{\rv} \\\hline
Max word length &16 \\\hline
Per-character encoding dim & 24 \\\hline
Activation &GeLU \\\hline
\# of projection layers &1 \\\hline
Projection layer dim &32 \\\hline
\# of fully-connected layers &2 \\\hline
Fully-connected layer dim & 256 \\\hline
Spatial dropout rate &0.0625 \\\hline
Dropout rate &0 \\\hline
Embedding activation &Tanh \\\hline
Embedding dim &256 \\\hline
Similarity dim &256 \\\hline
\end{tabular}

\caption{\rv model hyperparameter details.}
\label{tab:app:rvb}
\end{table}

\section{Benchmarking Models}
\label{app:cla:mod}

In this section, we provide more detailed model hyperparameters for the evaluation models used in Section \ref{sec:cla} and Section \ref{sec:large}.

\paragraph{RNN} We used a Stacked-LSTM architecture, with the following hyperparameters:
\begin{itemize}
\item Dim: 256
\item Layers: 4
\item Dropout rate: 0.1
\end{itemize}

\paragraph{DPCNN}
We use the architecture described in~\cite{johnsonDeepPyramidConvolutional2017}, with the following hyperparameters:
\begin{itemize}
\item Filters: 256
\item Layers: 6
\item Kernel size: 3
\item Final dropout: 0.5
\item Activation: ReLU
\end{itemize}

\paragraph{BERT-Mini}
We use the architecture as described in~\cite{devlinBERTPretrainingDeep2019} for BERT-Mini, with the following hyperparameters:
\begin{itemize}
\item Layers: 4
\item Hidden dim: 256
\item Intermediate dim: 1024
\item Self-attention heads: 4
\item Dropout rate: 0.1
\item Activation: GeLU
\end{itemize}

\paragraph{BERT-Base}
We use the architecture described in~\cite{devlinBERTPretrainingDeep2019} for BERT-Base, with the following hyperparameters:
\begin{itemize}
\item Layers: 12
\item Hidden dim: 768
\item Intermediate dim: 3072
\item Self-attention heads: 12
\item Dropout rate: 0.1
\item Activation: GeLU
\end{itemize}
\section{\rv Ablation Studies}
\label{app:abl}

In this section, we present our ablation study results for the \rv model design and hyperparameter selection. Results are reported on the Multilingual Amazon Reviews dataset following the methodology described in Section~\ref{sec:abl}.

\subsection{Embedding Dimension}

Detailed results on how \rv's embedding layer dimension affects classification performance are reported in Table~\ref{tab:app:abl:dim}.

\begin{table*}[htb]
    \centering
    \resizebox{0.74\textwidth}{!}{%
    \footnotesize
\begin{tabular}{|r|r|c|rrrr|}
\hline
\multicolumn{1}{|l|}{\multirow{2}{*}{\textbf{Embedding Dim}}} & \multicolumn{1}{c|}{\multirow{2}{*}{\textbf{Pre-training Loss}}} & \multirow{2}{*}{\textbf{\# Params}} & \multicolumn{4}{c|}{\textbf{Test Accuracy}} \\ \cline{4-7} 
\multicolumn{1}{|l|}{} & \multicolumn{1}{c|}{} &  & \multicolumn{1}{c|}{\textbf{RNN}} & \multicolumn{1}{c|}{\textbf{CNN}} & \multicolumn{1}{c|}{\textbf{BERT}} & \multicolumn{1}{c|}{\textbf{AVG}} \\ \hline
64 & 0.0284 & 181k & \multicolumn{1}{r|}{92.8\%} & \multicolumn{1}{r|}{91.8\%} & \multicolumn{1}{r|}{92.5\%} & 92.4\% \\ \hline
100 & 0.0267 & 190k & \multicolumn{1}{r|}{93.2\%} & \multicolumn{1}{r|}{92.1\%} & \multicolumn{1}{r|}{92.4\%} & 92.6\% \\ \hline
128 & 0.0260 & 197k & \multicolumn{1}{r|}{93.3\%} & \multicolumn{1}{r|}{92.2\%} & \multicolumn{1}{r|}{92.5\%} & 92.7\% \\ \hline
200 & 0.0253 & 216k & \multicolumn{1}{r|}{93.4\%} & \multicolumn{1}{r|}{92.3\%} & \multicolumn{1}{r|}{92.6\%} & 92.7\% \\ \hline
\textbf{256} & 0.0248 & 230k & \multicolumn{1}{r|}{93.6\%} & \multicolumn{1}{r|}{92.3\%} & \multicolumn{1}{r|}{92.8\%} & 92.9\% \\ \hline
300 & 0.0245 & 241k & \multicolumn{1}{r|}{93.6\%} & \multicolumn{1}{r|}{92.3\%} & \multicolumn{1}{r|}{92.6\%} & 92.8\% \\ \hline
384 & 0.0237 & 263k & \multicolumn{1}{r|}{93.6\%} & \multicolumn{1}{r|}{92.2\%} & \multicolumn{1}{r|}{92.6\%} & 92.8\% \\ \hline
512 & 0.0225 & 296k & \multicolumn{1}{r|}{93.7\%} & \multicolumn{1}{r|}{92.3\%} & \multicolumn{1}{r|}{92.8\%} & 92.9\% \\ \hline
\end{tabular}
}
\caption{Ablation study results on the effect of the \rv pre-trained model's embedding dimension on classification performance. \textbf{Bold} denotes the hyperparameter selected for the final RETVec model.}
\label{tab:app:abl:dim}
\end{table*}

\subsection{Model Architecture}

Detailed results on the effect of \rv architecture type on classification performance are reported in Table~\ref{tab:app:abl:arch}.

\begin{table*}[htb]
    \footnotesize
    \centering

\begin{tabular}{|r|r|rrrr|}
\hline
\multicolumn{1}{|l|}{} & \multicolumn{1}{l|}{} & \multicolumn{4}{c|}{\textbf{Test Accuracy}} \\ \cline{3-6} 
\multicolumn{1}{|l|}{\textbf{Architecture}} & \multicolumn{1}{c|}{\textbf{Pre-training Loss}} & \multicolumn{1}{c|}{\textbf{RNN}} & \multicolumn{1}{c|}{\textbf{CNN}} & \multicolumn{1}{c|}{\textbf{BERT}} & \multicolumn{1}{c|}{\textbf{AVG}} \\ \hline
\textbf{MLP} & 0.0248 & \multicolumn{1}{r|}{93.6\%} & \multicolumn{1}{r|}{92.3\%} & \multicolumn{1}{r|}{92.8\%} & 92.9\% \\ \hline
MLP + BERT & 0.0129 & \multicolumn{1}{r|}{93.2\%} & \multicolumn{1}{r|}{92.0\%} & \multicolumn{1}{r|}{92.4\%} & 92.5\% \\ \hline
MLP + T5 & 0.0120 & \multicolumn{1}{r|}{93.5\%} & \multicolumn{1}{r|}{92.2\%} & \multicolumn{1}{r|}{92.5\%} & 92.7\% \\ \hline
MLP + GAU & 0.0133 & \multicolumn{1}{r|}{93.5\%} & \multicolumn{1}{r|}{92.2\%} & \multicolumn{1}{r|}{92.8\%} & 92.8\% \\ \hline
MLP + LSTM & 0.0179 & \multicolumn{1}{r|}{93.4\%} & \multicolumn{1}{r|}{91.9\%} & \multicolumn{1}{r|}{92.5\%} & 92.6\% \\ \hline
MLP + CNN & 0.0214 & \multicolumn{1}{r|}{93.5\%} & \multicolumn{1}{r|}{92.1\%} & \multicolumn{1}{r|}{92.6\%} & 92.7\% \\ \hline
\end{tabular}

\caption{Ablation study results on \rv model architecture type on classification performance. \textbf{Bold} denotes the hyperparameter selected for the final RETVec model.}
\label{tab:app:abl:arch}
\end{table*}

\subsection{Maximum Word length}

Detailed results on how \rv's maximum input word length affects classification performance are reported in Table~\ref{tab:app:abl:len}.

\begin{table*}[htb]
    \centering
    \resizebox{0.7\textwidth}{!}{%
    \footnotesize

\begin{tabular}{|r|r|c|rrrr|}
\hline
\multicolumn{1}{|l|}{\multirow{2}{*}{\textbf{Word Len}}} & \multicolumn{1}{c|}{\multirow{2}{*}{\textbf{Pre-training Loss}}} & \multirow{2}{*}{\textbf{\# Params}} & \multicolumn{4}{c|}{\textbf{Test Accuracy}} \\ \cline{4-7} 
\multicolumn{1}{|l|}{} & \multicolumn{1}{c|}{} &  & \multicolumn{1}{c|}{\textbf{RNN}} & \multicolumn{1}{c|}{\textbf{CNN}} & \multicolumn{1}{c|}{\textbf{BERT}} & \multicolumn{1}{c|}{\textbf{AVG}} \\ \hline
8 & 0.0428 & 181k & \multicolumn{1}{r|}{93.6\%} & \multicolumn{1}{r|}{92.3\%} & \multicolumn{1}{r|}{92.7\%} & 92.8\% \\ \hline
10 & 0.0326 & 193k & \multicolumn{1}{r|}{93.5\%} & \multicolumn{1}{r|}{92.4\%} & \multicolumn{1}{r|}{92.6\%} & 92.8\% \\ \hline
12 & 0.0267 & 206k & \multicolumn{1}{r|}{93.5\%} & \multicolumn{1}{r|}{92.2\%} & \multicolumn{1}{r|}{92.7\%} & 92.8\% \\ \hline
14 & 0.0259 & 218k & \multicolumn{1}{r|}{93.5\%} & \multicolumn{1}{r|}{92.2\%} & \multicolumn{1}{r|}{92.6\%} & 92.8\% \\ \hline
\textbf{16} & 0.0248 & 230k & \multicolumn{1}{r|}{93.6\%} & \multicolumn{1}{r|}{92.3\%} & \multicolumn{1}{r|}{92.8\%} & 92.9\% \\ \hline
20 & 0.0242 & 255k & \multicolumn{1}{r|}{93.5\%} & \multicolumn{1}{r|}{92.2\%} & \multicolumn{1}{r|}{92.6\%} & 92.7\% \\ \hline
24 & 0.0234 & 280k & \multicolumn{1}{r|}{93.6\%} & \multicolumn{1}{r|}{92.3\%} & \multicolumn{1}{r|}{92.7\%} & 92.9\% \\ \hline
28 & 0.0246 & 304k & \multicolumn{1}{r|}{93.5\%} & \multicolumn{1}{r|}{92.2\%} & \multicolumn{1}{r|}{92.7\%} & 92.8\% \\ \hline
32 & 0.0235 & 328k & \multicolumn{1}{r|}{93.6\%} & \multicolumn{1}{r|}{92.2\%} & \multicolumn{1}{r|}{92.6\%} & 92.8\% \\ \hline
\end{tabular}
}
\caption{Ablation study results on the effect of maximum word length on \rv classification performance. \textbf{Bold} denotes the hyperparameter selected for the final RETVec model.}
\label{tab:app:abl:len}
\end{table*}

\subsection{Pre-training Loss Hyperparameters}

Detailed results on the effect of Multi-Similarity loss hyperparameters on \rv classification performance are reported in Table~\ref{tab:app:abl:multisim}. We also experimented with Circle Loss~\cite{sunCircleLossUnified2020} and report the results in Table~\ref{tab:app:abl:circle}.

\begin{table*}[h!]
    \centering
    \resizebox{0.75\textwidth}{!}{%
    \footnotesize
\begin{tabular}{|rrr|r|rrrr|}
\hline
\multicolumn{3}{|c|}{\textbf{Hyperparameter}} & \multicolumn{1}{l|}{} & \multicolumn{4}{c|}{\textbf{Test Accuracy}} \\ \hline
\multicolumn{1}{|c|}{\textbf{$\alpha$}} & \multicolumn{1}{c|}{\textbf{$\beta$}} & \multicolumn{1}{c|}{\textbf{$\lambda$}} & \multicolumn{1}{c|}{\textbf{Pre-training Loss}} & \multicolumn{1}{c|}{\textbf{RNN}} & \multicolumn{1}{c|}{\textbf{CNN}} & \multicolumn{1}{c|}{\textbf{BERT}} & \multicolumn{1}{c|}{\textbf{AVG}} \\ \hline
\multicolumn{1}{|r|}{2} & \multicolumn{1}{r|}{20} & 0.5 & 0.0432 & \multicolumn{1}{r|}{93.5\%} & \multicolumn{1}{r|}{92.1\%} & \multicolumn{1}{r|}{92.4\%} & 92.7\% \\ \hline
\multicolumn{1}{|r|}{2} & \multicolumn{1}{r|}{20} & 1.0 & 0.2643 & \multicolumn{1}{r|}{92.9\%} & \multicolumn{1}{r|}{91.2\%} & \multicolumn{1}{r|}{92.6\%} & 92.2\% \\ \hline
\multicolumn{1}{|r|}{2} & \multicolumn{1}{r|}{40} & 0.5 & 0.0455 & \multicolumn{1}{r|}{93.6\%} & \multicolumn{1}{r|}{92.1\%} & \multicolumn{1}{r|}{92.6\%} & 92.7\% \\ \hline
\multicolumn{1}{|r|}{2} & \multicolumn{1}{r|}{40} & 1.0 & 0.3610 & \multicolumn{1}{r|}{92.9\%} & \multicolumn{1}{r|}{91.2\%} & \multicolumn{1}{r|}{92.4\%} & 92.2\% \\ \hline
\multicolumn{1}{|r|}{2} & \multicolumn{1}{r|}{80} & 0.5 & 0.0464 & \multicolumn{1}{r|}{93.5\%} & \multicolumn{1}{r|}{92.3\%} & \multicolumn{1}{r|}{92.6\%} & 92.8\% \\ \hline
\multicolumn{1}{|r|}{2} & \multicolumn{1}{r|}{80} & 1.0 & 0.3583 & \multicolumn{1}{r|}{92.6\%} & \multicolumn{1}{r|}{91.1\%} & \multicolumn{1}{r|}{92.4\%} & 92.0\% \\ \hline
\multicolumn{1}{|r|}{4} & \multicolumn{1}{r|}{20} & 0.5 & 0.0270 & \multicolumn{1}{r|}{93.4\%} & \multicolumn{1}{r|}{92.0\%} & \multicolumn{1}{r|}{92.6\%} & 92.7\% \\ \hline
\multicolumn{1}{|r|}{4} & \multicolumn{1}{r|}{20} & 1.0 & 0.1537 & \multicolumn{1}{r|}{93.0\%} & \multicolumn{1}{r|}{91.3\%} & \multicolumn{1}{r|}{92.4\%} & 92.2\% \\ \hline
\multicolumn{1}{|r|}{4} & \multicolumn{1}{r|}{40} & 0.5 & 0.0242 & \multicolumn{1}{r|}{93.5\%} & \multicolumn{1}{r|}{92.4\%} & \multicolumn{1}{r|}{92.6\%} & 92.8\% \\ \hline
\multicolumn{1}{|r|}{4} & \multicolumn{1}{r|}{40} & 1.0 & 0.1919 & \multicolumn{1}{r|}{92.8\%} & \multicolumn{1}{r|}{91.3\%} & \multicolumn{1}{r|}{92.4\%} & 92.2\% \\ \hline
\multicolumn{1}{|r|}{\textbf{4}} & \multicolumn{1}{r|}{\textbf{80}} & \textbf{0.5} & 0.0248 & \multicolumn{1}{r|}{93.6\%} & \multicolumn{1}{r|}{92.3\%} & \multicolumn{1}{r|}{92.8\%} & 92.9\% \\ \hline
\multicolumn{1}{|r|}{4} & \multicolumn{1}{r|}{80} & 1.0 & 0.1851 & \multicolumn{1}{r|}{92.8\%} & \multicolumn{1}{r|}{91.2\%} & \multicolumn{1}{r|}{92.4\%} & 92.1\% \\ \hline
\end{tabular}
}
\caption{Ablation study results on the effect of Multi-Similarity loss hyperparameters on \rv classification performance. $\epsilon=0.1$ is fixed for all experiments. \textbf{Bold} denote the hyperparameters selected for the final RETVec model.}
\label{tab:app:abl:multisim}
\end{table*}

\begin{table*}[h!]
    \centering
    \resizebox{0.75\textwidth}{!}{%
    \footnotesize
\begin{tabular}{|rr|r|rrrr|}
\hline
\multicolumn{2}{|c|}{\textbf{Circle-Loss Hyperparameter}} & \multicolumn{1}{l|}{} & \multicolumn{4}{c|}{\textbf{Test Accuracy}} \\ \hline
\multicolumn{1}{|l|}{\textbf{Scale Factor $\gamma$}} & \multicolumn{1}{l|}{\textbf{Relaxation Factor $m$}} & \multicolumn{1}{c|}{\textbf{Pre-training Loss}} & \multicolumn{1}{c|}{\textbf{RNN}} & \multicolumn{1}{c|}{\textbf{CNN}} & \multicolumn{1}{c|}{\textbf{BERT}} & \multicolumn{1}{c|}{\textbf{AVG}} \\ \hline
\multicolumn{1}{|r|}{64} & 0.3 & 7.55 & \multicolumn{1}{r|}{93.3\%} & \multicolumn{1}{r|}{91.9\%} & \multicolumn{1}{r|}{92.6\%} & 92.6\% \\ \hline
\multicolumn{1}{|r|}{64} & 0.4 & 2.77 & \multicolumn{1}{r|}{93.5\%} & \multicolumn{1}{r|}{92.0\%} & \multicolumn{1}{r|}{92.4\%} & 92.7\% \\ \hline
\multicolumn{1}{|r|}{64} & 0.5 & 0.63 & \multicolumn{1}{r|}{93.6\%} & \multicolumn{1}{r|}{92.4\%} & \multicolumn{1}{r|}{92.6\%} & 92.9\% \\ \hline
\multicolumn{1}{|r|}{128} & 0.3 & 12.85 & \multicolumn{1}{r|}{93.3\%} & \multicolumn{1}{r|}{91.9\%} & \multicolumn{1}{r|}{92.6\%} & 92.6\% \\ \hline
\multicolumn{1}{|r|}{128} & 0.4 & 5.06 & \multicolumn{1}{r|}{93.6\%} & \multicolumn{1}{r|}{92.1\%} & \multicolumn{1}{r|}{92.6\%} & 92.8\% \\ \hline
\multicolumn{1}{|r|}{128} & 0.5 & 1.18 & \multicolumn{1}{r|}{93.6\%} & \multicolumn{1}{r|}{92.3\%} & \multicolumn{1}{r|}{92.6\%} & 92.9\% \\ \hline
\multicolumn{1}{|r|}{256} & 0.3 & 24.97 & \multicolumn{1}{r|}{93.3\%} & \multicolumn{1}{r|}{91.9\%} & \multicolumn{1}{r|}{92.5\%} & 92.5\% \\ \hline
\multicolumn{1}{|r|}{256} & 0.4 & 9.49 & \multicolumn{1}{r|}{93.5\%} & \multicolumn{1}{r|}{92.0\%} & \multicolumn{1}{r|}{92.4\%} & 92.7\% \\ \hline
\multicolumn{1}{|r|}{256} & 0.5 & 2.59 & \multicolumn{1}{r|}{93.7\%} & \multicolumn{1}{r|}{92.4\%} & \multicolumn{1}{r|}{92.5\%} & 92.9\% \\ \hline
\multicolumn{1}{|r|}{512} & 0.3 & 49.01 & \multicolumn{1}{r|}{93.2\%} & \multicolumn{1}{r|}{92.0\%} & \multicolumn{1}{r|}{92.6\%} & 92.6\% \\ \hline
\multicolumn{1}{|r|}{512} & 0.4 & 19.88 & \multicolumn{1}{r|}{93.3\%} & \multicolumn{1}{r|}{92.2\%} & \multicolumn{1}{r|}{92.7\%} & 92.7\% \\ \hline
\multicolumn{1}{|r|}{512} & 0.5 & 5.26 & \multicolumn{1}{r|}{93.6\%} & \multicolumn{1}{r|}{92.4\%} & \multicolumn{1}{r|}{92.7\%} & 92.9\% \\ \hline
\end{tabular}
}
\caption{Ablation study results for pre-training \rv with various Circle Loss~\cite{sunCircleLossUnified2020} hyperparameter settings.}
\label{tab:app:abl:circle}
\end{table*}

\subsection{Model Capacity}

Detailed results on how the number and dimension of fully-connected dense layers in the \rv model affects classification performance are presented in Table~\ref{tab:app:abl:capacity}.

\begin{table*}[h!]
    \centering
    \resizebox{0.75\textwidth}{!}{%
    \footnotesize
\begin{tabular}{|rr|c|r|rrrr|}
\hline
\multicolumn{2}{|c|}{\textbf{Model Capacity}} & \multirow{2}{*}{\textbf{\# Params}} & \multicolumn{1}{c|}{\multirow{2}{*}{\textbf{Pre-training Loss}}} & \multicolumn{4}{c|}{\textbf{Test Accuracy}} \\ \cline{1-2} \cline{5-8} 
\multicolumn{1}{|l|}{\textbf{Dense Layers}} & \multicolumn{1}{l|}{\textbf{Dense Layer Dim}} &  & \multicolumn{1}{c|}{} & \multicolumn{1}{c|}{\textbf{RNN}} & \multicolumn{1}{c|}{\textbf{CNN}} & \multicolumn{1}{c|}{\textbf{BERT}} & \multicolumn{1}{c|}{\textbf{AVG}} \\ \hline
\multicolumn{1}{|r|}{0} & - & 99k & 0.0445 & \multicolumn{1}{r|}{92.8\%} & \multicolumn{1}{r|}{91.6\%} & \multicolumn{1}{r|}{92.3\%} & 92.2\% \\ \hline
\multicolumn{1}{|r|}{1} & 128 & 82k & 0.0383 & \multicolumn{1}{r|}{93.5\%} & \multicolumn{1}{r|}{91.8\%} & \multicolumn{1}{r|}{92.4\%} & 92.6\% \\ \hline
\multicolumn{1}{|r|}{1} & 256 & 164k & 0.0312 & \multicolumn{1}{r|}{93.5\%} & \multicolumn{1}{r|}{91.8\%} & \multicolumn{1}{r|}{92.4\%} & 92.6\% \\ \hline
\multicolumn{1}{|r|}{1} & 384 & 246k & 0.0284 & \multicolumn{1}{r|}{93.6\%} & \multicolumn{1}{r|}{92.0\%} & \multicolumn{1}{r|}{92.6\%} & 92.7\% \\ \hline
\multicolumn{1}{|r|}{1} & 512 & 328k & 0.0258 & \multicolumn{1}{r|}{93.6\%} & \multicolumn{1}{r|}{92.2\%} & \multicolumn{1}{r|}{92.8\%} & 92.8\% \\ \hline
\multicolumn{1}{|r|}{2} & 128 & 989k & 0.0334 & \multicolumn{1}{r|}{93.4\%} & \multicolumn{1}{r|}{91.9\%} & \multicolumn{1}{r|}{92.6\%} & 92.6\% \\ \hline
\multicolumn{1}{|r|}{\textbf{2}} & \textbf{256} & 230k & 0.0248 & \multicolumn{1}{r|}{93.6\%} & \multicolumn{1}{r|}{92.3\%} & \multicolumn{1}{r|}{92.8\%} & 92.9\% \\ \hline
\multicolumn{1}{|r|}{2} & 384 & 394k & 0.0201 & \multicolumn{1}{r|}{93.5\%} & \multicolumn{1}{r|}{92.3\%} & \multicolumn{1}{r|}{92.6\%} & 92.8\% \\ \hline
\multicolumn{1}{|r|}{2} & 512 & 591k & 0.0177 & \multicolumn{1}{r|}{93.7\%} & \multicolumn{1}{r|}{92.3\%} & \multicolumn{1}{r|}{92.6\%} & 92.9\% \\ \hline
\multicolumn{1}{|r|}{3} & 128 & 115k & 0.0314 & \multicolumn{1}{r|}{93.4\%} & \multicolumn{1}{r|}{91.9\%} & \multicolumn{1}{r|}{92.5\%} & 92.6\% \\ \hline
\multicolumn{1}{|r|}{3} & 256 & 296k & 0.0213 & \multicolumn{1}{r|}{93.6\%} & \multicolumn{1}{r|}{92.4\%} & \multicolumn{1}{r|}{92.5\%} & 92.8\% \\ \hline
\multicolumn{1}{|r|}{3} & 384 & 542k & 0.0175 & \multicolumn{1}{r|}{93.6\%} & \multicolumn{1}{r|}{92.2\%} & \multicolumn{1}{r|}{92.7\%} & 92.8\% \\ \hline
\multicolumn{1}{|r|}{3} & 512 & 854k & 0.0149 & \multicolumn{1}{r|}{93.5\%} & \multicolumn{1}{r|}{92.4\%} & \multicolumn{1}{r|}{92.6\%} & 92.8\% \\ \hline
\end{tabular}
}
\caption{Ablation study results on the effect of \rv model capacity (number and dimension of the fully-connected layers) on classification performance. \textbf{Bold} denote the hyperparameters selected for the final RETVec model.}
\label{tab:app:abl:capacity}
\end{table*}

\subsection{Spatial Dropout Rate}

Detailed ablation study results on the amount of spatial dropout in the \rv model and its effect on classification performance are presented in Table~\ref{tab:app:abl:dropout}. Increments of $1/16$ were used because it corresponds to dropping out one character of the input on average, since \rv's model accepts an input of up to 16 characters per word.

\begin{table*}[h!]
    \centering
    \resizebox{0.7\textwidth}{!}{%
    \footnotesize
\begin{tabular}{|r|r|rrrr|}
\hline
\multicolumn{1}{|l|}{} & \multicolumn{1}{l|}{} & \multicolumn{4}{c|}{\textbf{Test Accuracy}} \\ \cline{3-6} 
\multicolumn{1}{|l|}{\textbf{Spatial Dropout}} & \multicolumn{1}{c|}{\textbf{Pre-training Loss}} & \multicolumn{1}{c|}{\textbf{RNN}} & \multicolumn{1}{c|}{\textbf{CNN}} & \multicolumn{1}{c|}{\textbf{BERT}} & \multicolumn{1}{c|}{\textbf{AVG}} \\ \hline
0.00\% & 0.0122 & \multicolumn{1}{r|}{93.4\%} & \multicolumn{1}{r|}{91.4\%} & \multicolumn{1}{r|}{92.5\%} & 92.4\% \\ \hline
\textbf{6.25\%} & 0.0248 & \multicolumn{1}{r|}{93.6\%} & \multicolumn{1}{r|}{92.3\%} & \multicolumn{1}{r|}{92.8\%} & 92.9\% \\ \hline
12.50\% & 0.0465 & \multicolumn{1}{r|}{93.4\%} & \multicolumn{1}{r|}{92.0\%} & \multicolumn{1}{r|}{92.3\%} & 92.6\% \\ \hline
18.75\% & 0.0722 & \multicolumn{1}{r|}{92.7\%} & \multicolumn{1}{r|}{91.5\%} & \multicolumn{1}{r|}{91.8\%} & 92.0\% \\ \hline
25.00\% & 0.0967 & \multicolumn{1}{r|}{92.7\%} & \multicolumn{1}{r|}{91.4\%} & \multicolumn{1}{r|}{91.2\%} & 91.8\% \\ \hline
\end{tabular}
}
\caption{Ablation study results on the effect of spatial dropout rate on the \rv input character encoding. \textbf{Bold} denotes the value selected for the final RETVec model.}
\label{tab:app:abl:dropout}
\end{table*}

\subsection{Pre-Training Objectives}

We evaluated combining \rv's pre-training objective (Multi-Similarity loss) with other objective functions and pre-training tasks as well. Specifically, we experimented with the following objectives: 1) augmentation position prediction, 2) augmentation position and type prediction, 3) decoding (predicting the character encoding of the input token), and 4) denoising (predicting the character encoding of the original, non-augmented token). Table~\ref{tab:app:abl:objective} reports the results of our experiments on different pre-training objectives.

\begin{table*}[h!]
    \centering
    \resizebox{\textwidth}{!}{%
    \footnotesize
\begin{tabular}{|l|r|r|r|r|r|r|}
\hline
\textbf{Objectives} & \multicolumn{1}{l|}{\textbf{Similarity Loss}} & \multicolumn{1}{c|}{\textbf{Total Loss}} & \multicolumn{1}{c|}{\textbf{RNN}} & \multicolumn{1}{c|}{\textbf{CNN}} & \multicolumn{1}{c|}{\textbf{BERT}} & \multicolumn{1}{c|}{\textbf{AVG}} \\ \hline
Similarity, Augmentation Position Detection & 0.0261 & 0.2228 & 93.4\% & 92.0\% & 92.8\% & 92.7\% \\ \hline
Similarity, Augmentation Position and Type Prediction & 0.0238 & 0.0970 & 93.2\% & 92.2\% & 92.6\% & 92.7\% \\ \hline
Similarity, Decoding & 0.0278 & 0.0384 & 93.5\% & 92.1\% & 92.6\% & 92.8\% \\ \hline
Similarity, Denoising & 0.0241 & 0.1088 & 93.3\% & 91.8\% & 92.7\% & 92.6\% \\ \hline
Similarity & 0.0248 & 0.0248 & 93.6\% & 92.3\% & 92.8\% & 92.9\% \\ \hline
\end{tabular}
}
\caption{Ablation study results on combining different pre-training objectives with similarity loss for \rv pre-training.}
\label{tab:app:abl:objective}
\end{table*}

\pagebreak

\section{\rv Pre-training Dataset Augmentations}
\label{app:aug}

% \begin{table*}[h!]
%     \centering
%     \resizebox{\textwidth}{!}{%
%     \footnotesize
% \begin{tabular}{|l|l|l|l|}
% \hline
% \multicolumn{1}{|l|}{\textbf{Insertion}} & \multicolumn{1}{l|}{\textbf{Substitution}} & \multicolumn{1}{l|}{\textbf{Transposition}} & \multicolumn{1}{l|}{\textbf{Deletion}} \\ \hline
% Repeated character insertion & Case substitution & Neighboring character swap & Character deletion \\
% n-grams based prefix insertion for n=3,4,5 & n-grams based substitution for n=3,4,5 & 3-character block random shuffle &  \\
% n-grams based suffix insertion for n=3,4,5 & QWERTY keyboard typo substitution &  &  \\
% Random ASCII character insertion & Homoglyphs substition &  &  \\
% Language alphabet-based random character insertion & Random ASCII character substitution &  &  \\
% Random punctuation insertion & Language alphabet-based random character subsitution &  &  \\
% Random punctuation prefix & Random puctuation substitution &  &  \\
% Random punctuation suffix & Random BMP Unicode substitution &  &  \\
% Random BMP Unicode insertion &  &  &  \\
% Random emoji prefix &  &  &  \\
% Random emoji suffix &  &  & 
% \end{tabular}
% }
% \caption{.}
% \label{tab:app:augment}
% \end{table*}

Below, we provide the full list of character-level augmentations (broken down into four categories) used to generate typo-augmented words for the \rv pre-training dataset, as described in Section~\ref{sec:pretrain}.

\begin{itemize}
    \item Deletion
    \item Insertion
    \begin{itemize}
    \item Repeated character insertion
    \item n-grams based prefix insertion for $n=3,4,5$
    \item n-grams based suffix insertion for $n=3,4,5$
    \item Random ASCII character insertion
    \item Language alphabet-based random character insertion
    \item Random punctuation insertion
    \item Random punctuation prefix
    \item Random punctuation suffix
    \item Random BMP Unicode insertion
    \item Random emoji prefix
    \item Random emoji suffix
    \end{itemize}
    \item Substitution
    \begin{itemize}
    \item Case substitution
    \item n-grams based substitution for $n=3,4,5$
    \item QWERTY keyboard typo substitution
    \item Homoglyphs substition
    \item Random ASCII character substitution
    \item Language alphabet-based random character subsitution
    \item Random puctuation substitution
    \item Random BMP Unicode substitution
    \end{itemize}
    \item Transposition
    \begin{itemize}
    \item Neighboring character swap
    \item 3-character block random shuffle
    \end{itemize}
\end{itemize}

% \begin{table}[htb]
%     \centering
%     \begin{tabular}{|l|r|}
%     \hline
% \footnotesize  
% Max Input Length &16 \\\hline
% \# of attention heads &1 \\\hline
% Attention kernel &$relu^2$ \\\hline
% Attention type &Quadratic \\\hline
% FFN type &GAU \\\hline
% Activation &Swish \\\hline
% Norm. type &ScaleNorm \\\hline
% Absolute position emb. &ScaledSin \\\hline
% Relative position emb. &RoPE \\\hline
% Number of layers &2 \\\hline
% Hidden size &32 \\\hline
% Expansion rate &1 \\\hline
% Dropout rate &0 \\\hline
% Embedding activation &Tanh \\\hline
% Embedding dim &256 \\\hline
% Similarity dim &256 \\\hline
%  Parameter count &148,451 \\\hline

% \end{tabular}

% \caption{\rvl model details.}
% \label{tab:rvl}
% \end{table}

% \begin{table}[h]
%     \centering
%     \begin{tabular}{|l|r|}
%     \hline
% \footnotesize  
% Max Input Length &16 \\\hline
% Activation &GeLU \\\hline
% Norm. type &Batch \\\hline
% Number of layers &2 \\\hline
% Hidden size &32 \\\hline
% Number of FC layers &1 \\\hline
% Dropout rate &0 \\\hline
% Embedding activation &Tanh \\\hline
% Embedding dim &256 \\\hline
% Similarity dim &256 \\\hline
%  Parameter count &201,536 \\\hline
% \end{tabular}

% \caption{\fix \rvb model details.}
% \label{tab:rvb}
% \end{table}

\section{\rv Pre-training Hyperparameters}
\label{app:tra}

We train \rv using Multi-Similarity loss with hyperparameters $\alpha=4$, $\beta=40$, $\epsilon=0.1$ and $\lambda=0.5$. Detailed pre-training hyperparameters are reported in Table~\ref{tab:tra}.

\begin{table}[h]
    \centering
    \begin{tabular}{|l|r|}
    \hline
\footnotesize  
\textbf{Hyperparameter} &\textbf{Pre-training} \\\hline
Training steps & 500k \\
Batch size &1024 \\
Adam $\epsilon$ &1.00e-7 \\
Adam $\beta_1$ &0.9 \\
Adam $\beta_2$ &0.999 \\
Weight decay &0 \\
Peak learning rate &0.001 \\
End learning rate &0.0001 \\
Warmup steps &10000 \\
Decay function &Cosine \\
\hline
\end{tabular}
\caption{\rv pre-training optimizer hyperparameters.}
\label{tab:tra}
\end{table}

\section{Typo Resilience Evaluation}
Detailed results for random mixed typo resilience across every dataset and vectorizer can be found in Table~\ref{app:tab:typofull}.

\begin{table*}[h]
    \resizebox{\textwidth}{!}{%
    \footnotesize
\begin{tabular}{|l|l|r|r|r|r|r|r|r|r|r|r|r|}
\hline
\textbf{Dataset} & \multicolumn{1}{l|}{\textbf{Vectorizer}} & \multicolumn{1}{r|}{\textbf{0\%}} & \multicolumn{1}{r|}{\textbf{10\%}} & \multicolumn{1}{r|}{\textbf{20\%}} & \multicolumn{1}{r|}{\textbf{30\%}} & \multicolumn{1}{r|}{\textbf{40\%}} & \multicolumn{1}{r|}{\textbf{50\%}} & \multicolumn{1}{r|}{\textbf{60\%}} & \multicolumn{1}{r|}{\textbf{70\%}} & \multicolumn{1}{r|}{\textbf{80\%}} & \multicolumn{1}{r|}{\textbf{90\%}} & \multicolumn{1}{r|}{\textbf{100\%}} \\ \hline
\multirow{6}{*}{AG News} & Whitespace & 92.0\% & 91.6\% & 91.0\% & 90.2\% & 88.6\% & 86.8\% & 83.8\% & 79.3\% & 72.3\% & 63.5\% & 49.1\% \\
 & SentencePiece & 91.8\% & 91.0\% & 89.9\% & 88.5\% & 86.2\% & 82.8\% & 78.7\% & 74.0\% & 67.9\% & 60.9\% & 51.5\% \\
 & BPE & 91.6\% & 90.5\% & 89.1\% & 87.6\% & 84.8\% & 80.6\% & 76.1\% & 70.2\% & 63.8\% & 56.2\% & 46.3\% \\
 & fastText & 92.8\% & 92.1\% & 91.6\% & 90.7\% & 89.1\% & 87.8\% & 85.1\% & 82.5\% & 77.2\% & 70.5\% & 59.7\% \\
  & RETVec-raw & 91.0\% & 90.3\% & 89.6\% & 88.3\% & 87.3\% & 85.5\% & 83.7\% & 81.8\% & 78.2\% & 74.7\% & 68.5\% \\
 & RETVec & 92.6\% & 92.1\% & 91.6\% & 91.1\% & 90.5\% & 89.9\% & 88.7\% & 87.3\% & 86.0\% & 84.2\% & 81.2\% \\ \hline
\multirow{6}{*}{Yelp P.} & Whitespace & 93.2\% & 92.5\% & 91.6\% & 90.5\% & 88.6\% & 86.7\% & 84.1\% & 80.4\% & 75.8\% & 69.7\% & 60.9\% \\
 & SentencePiece & 93.1\% & 91.5\% & 89.7\% & 87.7\% & 85.1\% & 81.9\% & 78.7\% & 75.1\% & 71.2\% & 67.4\% & 62.6\% \\
 & BPE & 93.0\% & 91.7\% & 90.2\% & 88.5\% & 86.2\% & 83.5\% & 80.5\% & 77.2\% & 73.1\% & 69.0\% & 64.2\% \\
 & fastText & 94.1\% & 93.5\% & 92.8\% & 92.0\% & 90.7\% & 89.3\% & 87.9\% & 85.8\% & 83.1\% & 80.2\% & 75.6\% \\ 
  & RETVec-raw & 92.4\% & 91.7\% & 90.9\% & 89.8\% & 88.6\% & 87.2\% & 85.7\% & 83.7\% & 81.2\% & 78.6\% & 74.5\% \\
 & RETVec & 92.7\% & 92.2\% & 91.6\% & 90.9\% & 90.0\% & 89.2\% & 88.0\% & 86.8\% & 85.1\% & 83.5\% & 80.9\% \\ \hline
\multirow{6}{*}{Multilingual Amazon P.}  & Whitespace & 92.7\% & 92.1\% & 91.6\% & 91.1\% & 90.2\% & 89.1\% & 87.9\% & 86.2\% & 83.7\% & 80.9\% & 75.0\% \\
 & SentencePiece & 89.6\% & 88.5\% & 87.5\% & 86.2\% & 84.5\% & 82.6\% & 80.7\% & 78.7\% & 76.1\% & 73.9\% & 70.7\% \\
 & BPE & 88.9\% & 87.7\% & 86.6\% & 85.4\% & 83.6\% & 81.8\% & 80.1\% & 77.9\% & 75.3\% & 73.4\% & 70.3\% \\
 & fastText & 86.2\% & 85.5\% & 84.8\% & 84.0\% & 82.9\% & 81.0\% & 80.1\% & 77.9\% & 75.9\% & 73.9\% & 70.2\% \\ 
  & RETVec-raw & 92.3\% & 91.6\% & 90.8\% & 90.0\% & 89.0\% & 87.7\% & 86.3\% & 84.8\% & 82.7\% & 80.6\% & 77.0\% \\
& RETVec & 92.9\% & 92.5\% & 92.0\% & 91.7\% & 91.3\% & 90.6\% & 90.2\% & 89.5\% & 88.6\% & 87.7\% & 86.1\% \\ \hline
\multirow{6}{*}{MASSIVE}  & Whitespace & 70.0\% & 58.9\% & 58.8\% & 58.4\% & 55.8\% & 51.7\% & 49.2\% & 43.9\% & 37.1\% & 34.8\% & 17.3\% \\
 & SentencePiece & 69.0\% & 58.5\% & 58.4\% & 57.8\% & 55.8\% & 52.7\% & 50.7\% & 46.7\% & 41.8\% & 40.9\% & 29.6\% \\
 & BPE & 65.5\% & 54.6\% & 54.4\% & 54.1\% & 52.0\% & 48.6\% & 46.9\% & 42.7\% & 38.4\% & 37.0\% & 26.4\% \\
 & fastText & 16.7\% & 15.5\% & 16.1\% & 15.2\% & 14.8\% & 14.3\% & 13.9\% & 13.7\% & 13.1\% & 13.0\% & 12.5\% \\ 
  & RETVec-raw & 69.6\% & 59.7\% & 59.7\% & 59.2\% & 57.4\% & 54.4\% & 52.7\% & 49.0\% & 44.8\% & 43.6\% & 32.5\% \\
 & RETVec & 73.2\% & 65.7\% & 65.7\% & 65.5\% & 63.9\% & 61.8\% & 60.4\% & 57.5\% & 54.1\% & 52.9\% & 43.5\% \\ \hline
\end{tabular}
    }
    \caption{Random mixed typo resilience results (0\% to 100\% word typo rate) for each classification dataset and vectorizer. Following the methodology described in Section~\ref{sec:typ}, test accuracy on each dataset is reported and results are averaged across the three model architectures we benchmarked in~\ref{sec:cla}.}
    \label{app:tab:typofull}
    \end{table*}

\section{Adversarial Resilience Evaluation}
\label{app:adv}

We report adversarial attack resilience results for all vectorizers, classification models, and adversarial attack algorithms we benchmarked in Table~\ref{tab:app:advfull}. The TextAttack~\cite{morrisTextAttackFrameworkAdversarial2020} framework was used to conduct all three types of adversarial attacks.

\begin{table*}[h]
    \resizebox{\textwidth}{!}{%
    \footnotesize
\begin{tabular}{|l|l|rrr|rrr|rrr|}
\hline
\multirow{2}{*}{\textbf{Model}} & \multirow{2}{*}{\textbf{Vectorizer}} & \multicolumn{3}{c|}{\textbf{TextBugger}} & \multicolumn{3}{c|}{\textbf{Pruthi}} & \multicolumn{3}{c|}{\textbf{DeepWordBug}} \\ \cline{3-11} 
 &  & \multicolumn{1}{l|}{\textbf{Original Acc}} & \multicolumn{1}{l|}{\textbf{Acc under Atk}} & \multicolumn{1}{l|}{\textbf{Atk Success \%}} & \multicolumn{1}{l|}{\textbf{Original Acc}} & \multicolumn{1}{l|}{\textbf{Acc under Atk}} & \multicolumn{1}{l|}{\textbf{Atk Success \%}} & \multicolumn{1}{l|}{\textbf{Original Acc}} & \multicolumn{1}{l|}{\textbf{Acc under Atk}} & \multicolumn{1}{l|}{\textbf{Atk Success \%}} \\ \hline
\multirow{6}{*}{LSTM} & Whitespace & \multicolumn{1}{r|}{90.6\%} & \multicolumn{1}{r|}{9.9\%} & 89.1\% & \multicolumn{1}{r|}{90.6\%} & \multicolumn{1}{r|}{84.2\%} & 7.1\% & \multicolumn{1}{r|}{90.6\%} & \multicolumn{1}{r|}{9.9\%} & 89.1\% \\ \cline{2-11} 
 & SentencePiece & \multicolumn{1}{r|}{90.3\%} & \multicolumn{1}{r|}{0.8\%} & 99.1\% & \multicolumn{1}{r|}{90.3\%} & \multicolumn{1}{r|}{68.1\%} & 24.6\% & \multicolumn{1}{r|}{90.3\%} & \multicolumn{1}{r|}{0.8\%} & 99.1\% \\ \cline{2-11} 
 & BPE & \multicolumn{1}{r|}{88.1\%} & \multicolumn{1}{r|}{3.3\%} & 96.3\% & \multicolumn{1}{r|}{88.1\%} & \multicolumn{1}{r|}{73.3\%} & 16.8\% & \multicolumn{1}{r|}{88.1\%} & \multicolumn{1}{r|}{3.3\%} & 96.3\% \\ \cline{2-11} 
 & fastText & \multicolumn{1}{r|}{92.7\%} & \multicolumn{1}{r|}{14.4\%} & 84.5\% & \multicolumn{1}{r|}{92.7\%} & \multicolumn{1}{r|}{83.3\%} & 10.1\% & \multicolumn{1}{r|}{92.7\%} & \multicolumn{1}{r|}{14.4\%} & 84.5\% \\ \cline{2-11} 
 & RETVec-raw & \multicolumn{1}{r|}{90.8\%} & \multicolumn{1}{r|}{22.3\%} & 75.4\% & \multicolumn{1}{r|}{90.8\%} & \multicolumn{1}{r|}{74.8\%} & 17.6\% & \multicolumn{1}{r|}{90.8\%} & \multicolumn{1}{r|}{22.3\%} & 75.4\% \\ \cline{2-11} 
 & RETVec & \multicolumn{1}{r|}{91.8\%} & \multicolumn{1}{r|}{23.7\%} & 74.2\% & \multicolumn{1}{r|}{91.8\%} & \multicolumn{1}{r|}{80.9\%} & 11.9\% & \multicolumn{1}{r|}{91.8\%} & \multicolumn{1}{r|}{23.7\%} & 74.2\% \\ \hline
\multirow{6}{*}{CNN} & Whitespace & \multicolumn{1}{r|}{90.9\%} & \multicolumn{1}{r|}{17.6\%} & 80.6\% & \multicolumn{1}{r|}{90.9\%} & \multicolumn{1}{r|}{83.8\%} & 7.8\% & \multicolumn{1}{r|}{90.9\%} & \multicolumn{1}{r|}{9.0\%} & 90.1\% \\ \cline{2-11} 
 & SentencePiece & \multicolumn{1}{r|}{90.3\%} & \multicolumn{1}{r|}{2.9\%} & 96.8\% & \multicolumn{1}{r|}{90.3\%} & \multicolumn{1}{r|}{72.2\%} & 20.0\% & \multicolumn{1}{r|}{90.3\%} & \multicolumn{1}{r|}{3.5\%} & 96.1\% \\ \cline{2-11} 
 & BPE & \multicolumn{1}{r|}{89.3\%} & \multicolumn{1}{r|}{31.8\%} & 64.4\% & \multicolumn{1}{r|}{89.3\%} & \multicolumn{1}{r|}{54.1\%} & 39.4\% & \multicolumn{1}{r|}{89.3\%} & \multicolumn{1}{r|}{43.5\%} & 51.3\% \\ \cline{2-11} 
 & fastText & \multicolumn{1}{r|}{91.9\%} & \multicolumn{1}{r|}{17.3\%} & 81.2\% & \multicolumn{1}{r|}{91.9\%} & \multicolumn{1}{r|}{74.9\%} & 18.5\% & \multicolumn{1}{r|}{91.9\%} & \multicolumn{1}{r|}{15.4\%} & 83.2\% \\ \cline{2-11} 
 & RETVec-raw & \multicolumn{1}{r|}{86.9\%} & \multicolumn{1}{r|}{30.1\%} & 65.4\% & \multicolumn{1}{r|}{86.9\%} & \multicolumn{1}{r|}{59.6\%} & 31.4\% & \multicolumn{1}{r|}{86.9\%} & \multicolumn{1}{r|}{37.7\%} & 56.6\% \\ \cline{2-11} 
 & RETVec & \multicolumn{1}{r|}{91.4\%} & \multicolumn{1}{r|}{34.3\%} & 62.5\% & \multicolumn{1}{r|}{91.4\%} & \multicolumn{1}{r|}{77.4\%} & 15.3\% & \multicolumn{1}{r|}{91.4\%} & \multicolumn{1}{r|}{45.0\%} & 50.8\% \\ \hline
\multirow{6}{*}{BERT} & Whitespace & \multicolumn{1}{r|}{89.4\%} & \multicolumn{1}{r|}{9.8\%} & 89.0\% & \multicolumn{1}{r|}{89.4\%} & \multicolumn{1}{r|}{83.6\%} & 6.5\% & \multicolumn{1}{r|}{89.4\%} & \multicolumn{1}{r|}{3.3\%} & 96.3\% \\ \cline{2-11} 
 & SentencePiece & \multicolumn{1}{r|}{89.8\%} & \multicolumn{1}{r|}{2.8\%} & 96.9\% & \multicolumn{1}{r|}{89.8\%} & \multicolumn{1}{r|}{70.8\%} & 21.2\% & \multicolumn{1}{r|}{89.8\%} & \multicolumn{1}{r|}{3.7\%} & 95.9\% \\ \cline{2-11} 
 & BPE & \multicolumn{1}{r|}{90.8\%} & \multicolumn{1}{r|}{8.2\%} & 91.0\% & \multicolumn{1}{r|}{90.8\%} & \multicolumn{1}{r|}{78.2\%} & 13.9\% & \multicolumn{1}{r|}{90.8\%} & \multicolumn{1}{r|}{1.2\%} & 98.7\% \\ \cline{2-11} 
 & fastText & \multicolumn{1}{r|}{92.6\%} & \multicolumn{1}{r|}{22.9\%} & 75.3\% & \multicolumn{1}{r|}{92.6\%} & \multicolumn{1}{r|}{80.5\%} & 13.1\% & \multicolumn{1}{r|}{92.6\%} & \multicolumn{1}{r|}{18.1\%} & 80.5\% \\ \cline{2-11} 
 & RETVec-raw & \multicolumn{1}{r|}{93.0\%} & \multicolumn{1}{r|}{30.8\%} & 66.9\% & \multicolumn{1}{r|}{93.0\%} & \multicolumn{1}{r|}{82.2\%} & 11.6\% & \multicolumn{1}{r|}{93.0\%} & \multicolumn{1}{r|}{38.9\%} & 58.2\% \\ \cline{2-11} 
 & RETVec & \multicolumn{1}{r|}{93.7\%} & \multicolumn{1}{r|}{30.1\%} & 67.9\% & \multicolumn{1}{r|}{93.7\%} & \multicolumn{1}{r|}{84.6\%} & 9.7\% & \multicolumn{1}{r|}{93.7\%} & \multicolumn{1}{r|}{40.5\%} & 56.8\% \\ \hline
\end{tabular}

    }
    \caption{Detailed adversarial resilience results on AG News. Results are reported on the same randomly selected 1000 examples from the AG News test split, following the methodology described in Section~\ref{sec:adv}.}
    \label{tab:app:advfull}
    \end{table*}

% \section{Pre-training BERT Evaluation}
% \label{app:bert}

\section{Pre-training and Fine-tuning BERT}
\label{app:pretrain_bert}

Table~\ref{tab:app:bert_hyper} details the hyperparameter settings used for pre-training and fine-tuning BERT-Base models. Table~\ref{tab:app:glue} shows detailed results on the GLUE benchmark, including the models' average performance and standard deviation for each GLUE task.

\begin{table*}[h]
\centering
\begin{tabular}{|c|c|c|}
\hline
\textbf{Hyperparameter} & \multicolumn{1}{c|}{\textbf{Pre-training}} & \multicolumn{1}{c|}{\textbf{Fine-tuning}} \\ \hline
Training steps & \multicolumn{1}{c|}{100k steps} & \multicolumn{1}{c|}{20 epochs} \\ 
Batch size & 64 & 32 \\ 
Sequence length & 512 & 512 \\ 
Adam $\epsilon$ & 1e-8 & 1e-8 \\ 
Adam $\beta_1$ & 0.9 & 0.9 \\ 
Adam $\beta_2$ & 0.999 & 0.999 \\ 
Weight decay & 0.01 & 0.01 \\ 
Max learning rate & 5e-5 & 2e-5 \\ 
End learning rate & 0 & 0 \\ 
Warmup steps & 10000 & \multicolumn{1}{c|}{First 5\% of steps} \\ 
Decay function & \multicolumn{1}{c|}{Linear} & \multicolumn{1}{c|}{None} \\ \hline
\end{tabular}
\caption{Pre-training and fine-tuning hyperparameters for BERT-Base models described in Section~\ref{sec:large}.}
\label{tab:app:bert_hyper}
\end{table*}

\begin{table*}[h]
    \resizebox{\textwidth}{!}{%
\begin{tabular}{|l|l|l|l|l|l|l|l|l|l|}
\hline
\textbf{Vectorizer} & \multicolumn{1}{l|}{\textbf{MNLI}} & \multicolumn{1}{l|}{\textbf{QNLI}} & \multicolumn{1}{l|}{\textbf{QQP}} & \multicolumn{1}{l|}{\textbf{RTE}} & \multicolumn{1}{l|}{\textbf{SST-2}} & \multicolumn{1}{l|}{\textbf{MRPC}} & \multicolumn{1}{l|}{\textbf{CoLA}} & \multicolumn{1}{l|}{\textbf{STS-B}} & \textbf{GLUE Avg} \\ \hline
SentencePiece & 80.6 (0.1) & \multicolumn{1}{r}{88.4 (0.3)} & 90.1 (0.0) & \textbf{66.1 (1.0)} & \underline{ 90.8 (0.3)} & 85.4 (0.l5) & \textbf{50.3 (0.4)} & \textbf{82.0 (0.5)} & \textbf{79.2 (0.3)} \\ \hline
\rvr & \textbf{82.0 (0.5)} & \textbf{89.5 (0.1)} & \textbf{90.4 (0.1)} & 64.5 (0.9) & \textbf{91.5 (0.1)} & \underline{ 86.3 (0.9)} & \underline{ 47.9 (1.1)} & 79.1 (0.2) & \underline{ 78.9 (0.4)} \\ \hline
\rvb & \underline{ 80.9 (0.4)} & \underline{ 88.9 (0.3)} & \underline{ 90.4 (0.1)} & \underline{ 65.0 (0.7)} & 90.7 (0.2) & \textbf{86.9 (0.4)} & 47.2 (0.7) & \underline{ 79.6 (0.2)} & 78.7 (0.3) \\ \hline
\end{tabular}
}
\caption{Detailed results on GLUE Benchmark for pre-trained BERT-Base models using \rv compared to SentencePiece. Each model is trained three times with different seeds, and the average and standard deviation is reported here. \textbf{Bold} indicates best results, \underline{underline} indicates second best.}
\label{tab:app:glue}
\end{table*}

\section{fastText Word Dataset}
\label{app:fasttext}

Table~\ref{tab:app:fasttext} contains statistics on word length computed on the fastText word dataset using words from all 157 available languages.

\begin{table}[h]
\centering
    \resizebox{0.5\textwidth}{!}{%
    \begin{tabular}{|*{7}{p{.9cm}}|}
    \hline
\textbf{Avg}&
\textbf{Median}&
\textbf{Std}&
\textbf{p90}&
\textbf{p95}&
\textbf{p99}&
\textbf{p99.9}\\\hline
8.4&
7.9&
4.6&
13.0&
\textbf{15.0}&
20.8&
\textbf{36.1}\\
\hline
\end{tabular}
}
\caption{Word length statistics computed on all fastText words from 157 languages. p90 denotes the 90th percentile, p95 denotes the 95th percentile, and so on.}
\label{tab:app:fasttext}
\end{table}

\section{Embedding Visualization}
\label{app:emb_viz}

We use the TensorBoard Embedding Projector to visualize \rv embeddings for words, as shown in Figure~\ref{app:emb_viz}. 

\begin{figure}[h]
\centering{
\includegraphics[width=\textwidth]{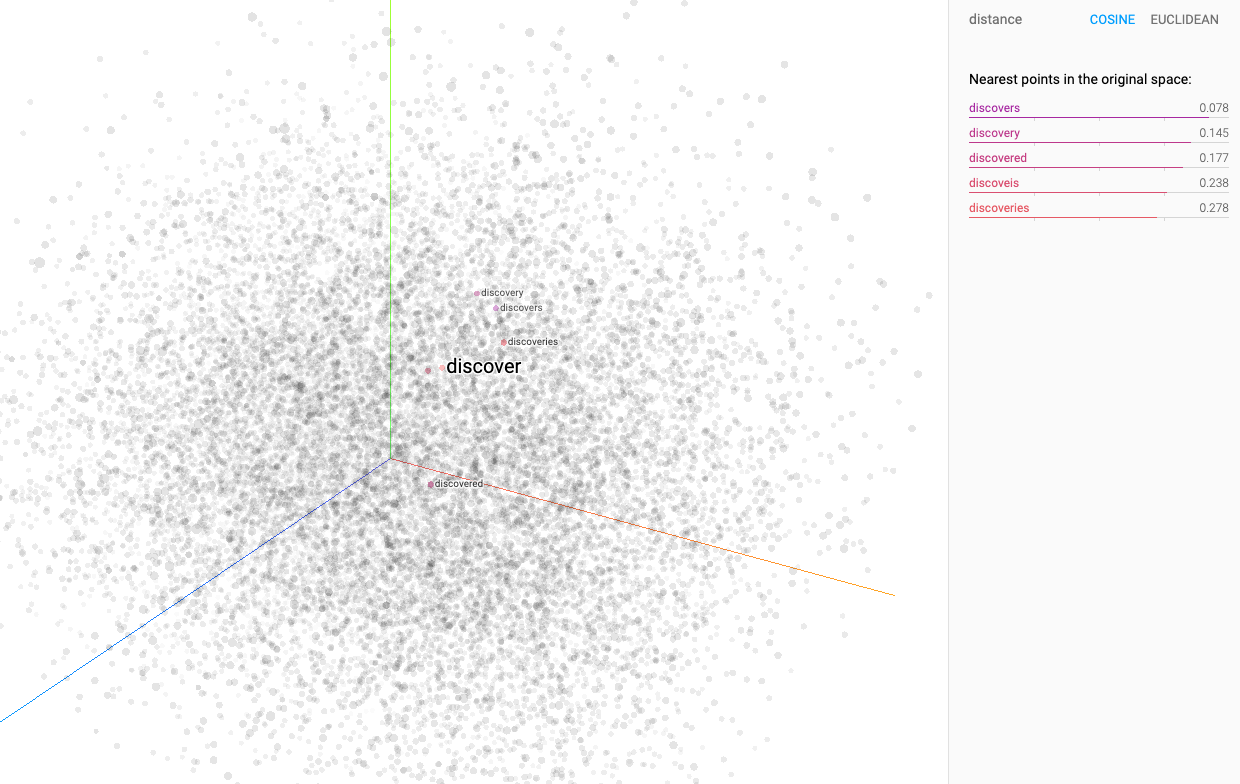}
\caption{\rv embedding visualization (using the TensorBoard Embedding Projector) of the most common 10000 English words and a typo-laden version of each word. The word 'discover' is selected as an example, and the 5 nearest neighbors and their cosine distances are shown.}
\label{fig:emb_viz}}
\end{figure}

%%%%%%%%%%%%%%%%%%%%%%%%%%%%%%%%%%%%%%%%%%%%%%%%%%%%%%%%%%%%%%%%%%%%%%%%%%%%%%%
%%%%%%%%%%%%%%%%%%%%%%%%%%%%%%%%%%%%%%%%%%%%%%%%%%%%%%%%%%%%%%%%%%%%%%%%%%%%%%%

%%%%%%%%%%%%%%%%%%%%%%%%%%%%%%%%%%%%%%%%%%%%%%%%%%%%%%%%%%%%

\end{document}